\documentclass[letterpaper, 10 pt, conference]{ieeeconf}
\IEEEoverridecommandlockouts
\usepackage{fancyhdr}
\usepackage[space]{cite}
\usepackage[font=small,skip=5pt]{caption}
\usepackage{amsmath,amsopn,amstext,amsfonts}
\usepackage{amssymb}
\usepackage{float}
\usepackage{siunitx}
\usepackage{graphicx}
\usepackage{textcomp}
\usepackage{xcolor}
\usepackage{multirow}
\usepackage{booktabs}
\usepackage{svg}
\usepackage{epstopdf}
\usepackage[linesnumbered,boxed,ruled,commentsnumbered]{algorithm2e}
\usepackage{algpseudocode}
\usepackage{geometry}
\def\BibTeX{{\rm B\kern-.05em{\sc i\kern-.025em b}\kern-.08em
    T\kern-.1667em\lower.7ex\hbox{E}\kern-.125emX}}
\begin{document}

\title{Pixel-level Extrinsic Self Calibration of High Resolution LiDAR and Camera in Targetless Environments}

\author{Chongjian Yuan, Xiyuan Liu, Xiaoping Hong, and Fu Zhang
\thanks{C. Yuan, X. Liu and F. Zhang are with the Department of Mechanical Engineering, The University of Hong Kong, Hong Kong Special Administrative Region, People's Republic of China.
{\tt\footnotesize $\{$ycj1,xliuaa$\}$@connect.hku.hk}, {\tt\footnotesize $ $fuzhang$ $@hku.hk}}
\thanks{X. Hong is with the School of System Design and Intelligent Manufacturing, Southern University of Science and Technology, Shenzhen, People’s Republic of China. {\tt\footnotesize $ $hongxp$ $@sustech.edu.cn}} \vspace{-0.8cm}
}
\maketitle

\begin{abstract}
In this letter, we present a novel method for automatic extrinsic calibration of high-resolution LiDARs and RGB cameras in targetless environments. Our approach does not require checkerboards but can achieve pixel-level accuracy by aligning natural edge features in the two sensors. On the theory level, we analyze the constraints imposed by edge features and the sensitivity of calibration accuracy with respect to edge distribution in the scene. On the implementation level, we carefully investigate the physical measuring principles of LiDARs and propose an efficient and accurate LiDAR edge extraction method based on point cloud voxel cutting and plane fitting. Due to the edges’ richness in natural scenes, we have carried out experiments in many indoor and outdoor scenes. The results show that this method has high robustness, accuracy, and consistency. It can promote the research and application of the fusion between LiDAR and camera. We have open sourced our code on GitHub\footnote{\label{github}https://github.com/hku-mars/livox\_camera\_calib} to benefit the community.
\end{abstract}


\section{Introduction}

Light detection and ranging (LiDAR) and camera sensors are commonly combined in developing autonomous driving vehicles. LiDAR sensor, owing to its direct 3D measurement capability, has been extensively applied to obstacle detection~\cite{xiao2018hybrid}, tracking~\cite{di2019behavioral}, and mapping~\cite{loamlivox} applications. The integrated onboard camera could also provide rich color information and facilitate the various LiDAR applications. With the recent rapid growing resolutions of LiDAR sensors, the demand for accurate extrinsic parameters becomes essential, especially for applications such as dense point cloud mapping, colorization, \textcolor{black}{and accurate and automated 3D surveying}. In this letter, our work deals with the accurate extrinsic calibration of high-resolution LiDAR and camera sensors.

Current extrinsic calibration methods rely heavily on external targets, such as checkerboard~\cite{ACSC,koo2020analytic,zhou2018automatic} or specific image pattern~\cite{chen2020novel}. By detecting, extracting, and matching feature points from both image and point cloud, the original problem is transformed into and solved with least-square equations. Due to its repetitive scanning pattern and the inevitable vibration of mechanical spinning LiDAR, e.g., Velodyne\footnote{https://velodynelidar.com/}, the reflected point cloud tends to be sparse and of large noise. This characteristic may mislead the cost function to the unstable results. Solid-state LiDAR, e.g., Livox~\cite{livox}, could compensate for this drawback with its dense point cloud. However, since these calibration targets are typically placed close to the sensor suite, the extrinsic error might be enlarged in the long-range scenario, e.g., large-scale point cloud colorization. In addition, it is not always practical to calibrate the sensors with targets being prepared at the start of each mission.

\begin{figure}
    \centering
    \includegraphics[width=1\linewidth]{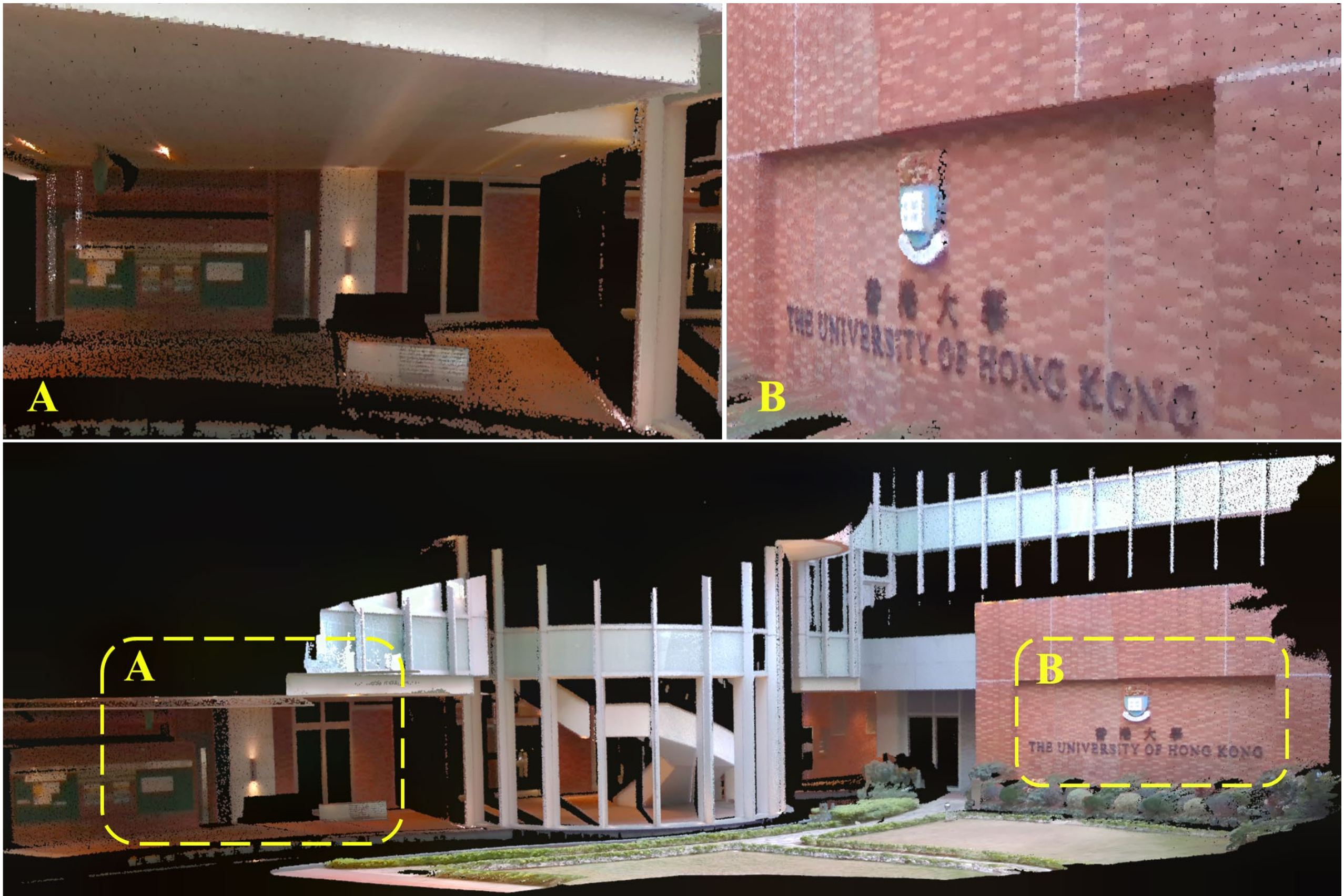}
    \caption{Point cloud of three aligned LiDAR scans colored using the proposed method. The accompanying experiment video has been uploaded to \textcolor{black}{https://youtu.be/e6Vkkasc4JI}.}
    \label{fig:color_pc}
    \vspace{-0.6cm}
\end{figure}

To address the above challenges, in this letter, we propose an automatic pixel-level extrinsic calibration method in targetless environments. This system operates by extracting natural edge features from the image and point cloud and minimizing the reprojection error. The proposed method does not rely on external targets, e.g., checkerboard, and is capable of functioning in both indoor and outdoor scenes. \textcolor{black}{Such a simple and convenient calibration allows one to calibrate the extrinsic parameters before or in the middle of each data collection or detect any misalignment of the sensors during their online operation that is usually not feasible with the target-based methods}. Specifically, our contributions are as follows:
\begin{itemize}
    \item We carefully study the underlying LiDAR measuring principle, which reveals that the commonly used depth-discontinuous edge features are not accurate nor reliable for calibration. \textcolor{black}{We propose a novel and reliable depth-continuous edge extraction algorithm that leads to more accurate calibration parameters.}
    \item We evaluate the robustness, consistency, and accuracy of our methods and implementation in various indoor and outdoor environments and compare our methods with other state-of-the-art. Results show that our method is robust to initial conditions, consistent to calibration scenes, and achieves pixel-level calibration accuracy in natural environments. Our method has an accuracy that is on par to (and sometimes even better than) target-based methods \textcolor{black}{and is applicable to both emerging solid-state and conventional spinning LiDARs.}
    \item Based on the analysis, we develop a practical calibration software and open source it on GitHub to benefit the community.
\end{itemize}

\section{Related Works}

Extrinsic calibration is a well-studied problem in robotics and is mainly divided into two categories: target-based and targetless. The primary distinction between them is how they define and extract features from both sensors. Geometric solids~\cite{kummerle2020,gong2013,park2014} and checkerboards~\cite{ACSC,koo2020analytic,zhou2018automatic} have been widely applied in target-based methods, due to its explicit constraints on plane normals and simplicity in problem formulation. As they require extra preparation, they are not practical, especially when they need to operate in a dynamically changing environment.

Targetless methods do not detect explicit geometric shapes from known targets. Instead, they use the more general plane and edge features that existed in nature. In~\cite{zhu2020camvox}, the LiDAR points are first projected onto the image plane and colored by depth and reflectivity values. Then 2D edges are extracted from this colormap and matched with those obtained from the image. Similarly, authors in~\cite{pandey2012automatic} optimize the extrinsic \textcolor{black}{calibration} by maximizing the mutual information between the colormap and the image. In~\cite{scaramuzza2007extrinsic,levinson2013automatic}, both authors detect and extract 3D edges from the point cloud by laser beam depth discontinuity. Then the 3D edges are back-projected onto the 2D image plane to calculate the residuals. The accuracy of edge estimation limits this method as the laser points do not strictly fall on the depth discontinuity margin. Motion-based methods have also been introduced in~\cite{nagy2019,lidarcalib} that the extrinsic is estimated from sensors' motion and refined by appearance information. \textcolor{black}{This motion-based calibration typically requires the sensor to move along a sufficient excited trajectory~\cite{lidarcalib}}. 

Our proposed method is a targetless method. Compared to \cite{pandey2012automatic, zhu2020camvox}, we directly extract 3D edge features in the point cloud, which suffer from no occlusion problem. Compared to \cite{scaramuzza2007extrinsic,levinson2013automatic}, we use depth-continuous edges, which proved to be more accurate and reliable. Our method works for a single pair LiDAR scan and achieves calibration accuracy comparable to target-based methods \cite{ACSC, zhou2018automatic}. 

\section{Methodology}\label{sec:methodology}
\subsection{Overview}\label{sec:overview}

\begin{figure}
    \centering
    \includegraphics[width=0.7\linewidth]{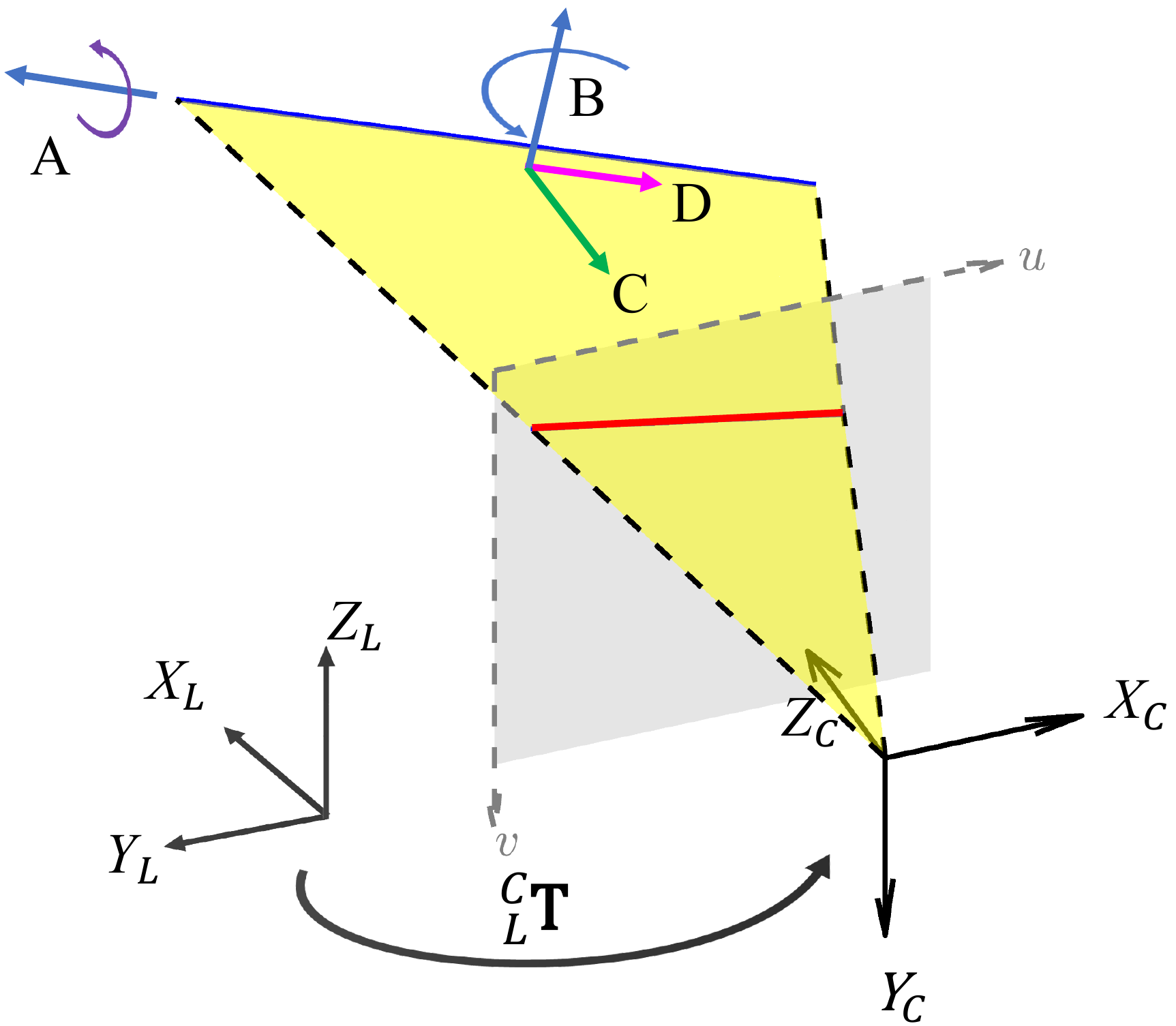}
    \caption{\textcolor{black}{Constraints imposed by an edge feature. The blue line represents the 3D edge, its projection to the image plane (the gray plane) produces the camera measurement (the red line). The edge after a translation along the axis C or axis D, or a rotation about the axis A (i.e., the edge itself) or axis B (except when the edge passes through the camera origin), remains on the same yellow plane and hence has the same projection on the image plane. That means, these four pose transformations on the edge (or equivalently on ${}^C_L \mathbf T$) are not distinguishable. }}
    \label{fig:line_constraints}
    \vspace{-0.6cm}
\end{figure}

Fig. \ref{fig:line_constraints} defines the coordinate frames involved in this paper: the LiDAR frame $L$, the camera frame $C$, and the 2D coordinate frame in the image plane. Denote ${}^C_L \mathbf T = ({}^C_L \mathbf R, {}^C_L \mathbf t) \in SE(3)$ the extrinsic between LiDAR and camera to be calibrated. Due to the wide availability of edge features in natural indoor and outdoor scenes, our method aligns these edge features observed by both LiDAR and camera sensors. 

Fig. \ref{fig:line_constraints} further illustrates the number of constraints imposed by a single edge to the extrinsic. As can be seen, the following degree of freedom (DoF) of the LiDAR pose relative to the camera cannot be distinguished: (1) translation along the edge (the red arrow D, Fig. \ref{fig:line_constraints}), (2) translation perpendicular to the edge (the green arrow C, Fig. \ref{fig:line_constraints}), (3) rotation about the normal vector of the plane formed by the edge and the camera focal point (the blue arrow B, Fig. \ref{fig:line_constraints}), and (4) rotation about the edge itself (the purple arrow A, Fig. \ref{fig:line_constraints}). As a result, a single edge feature constitutes two effective constraints to the extrinsic ${}^C_L \mathbf T$.  To obtain sufficient constraints to the extrinsic, we extract edge features of different orientations and locations, as detailed in the following section. 

\subsection{Edge Extraction and Matching}\label{sec:edge}
\subsubsection{Edge Extraction}

\begin{figure}[t]
    \vspace{-0.3cm}
    \centering
    \includegraphics[width=0.9\linewidth]{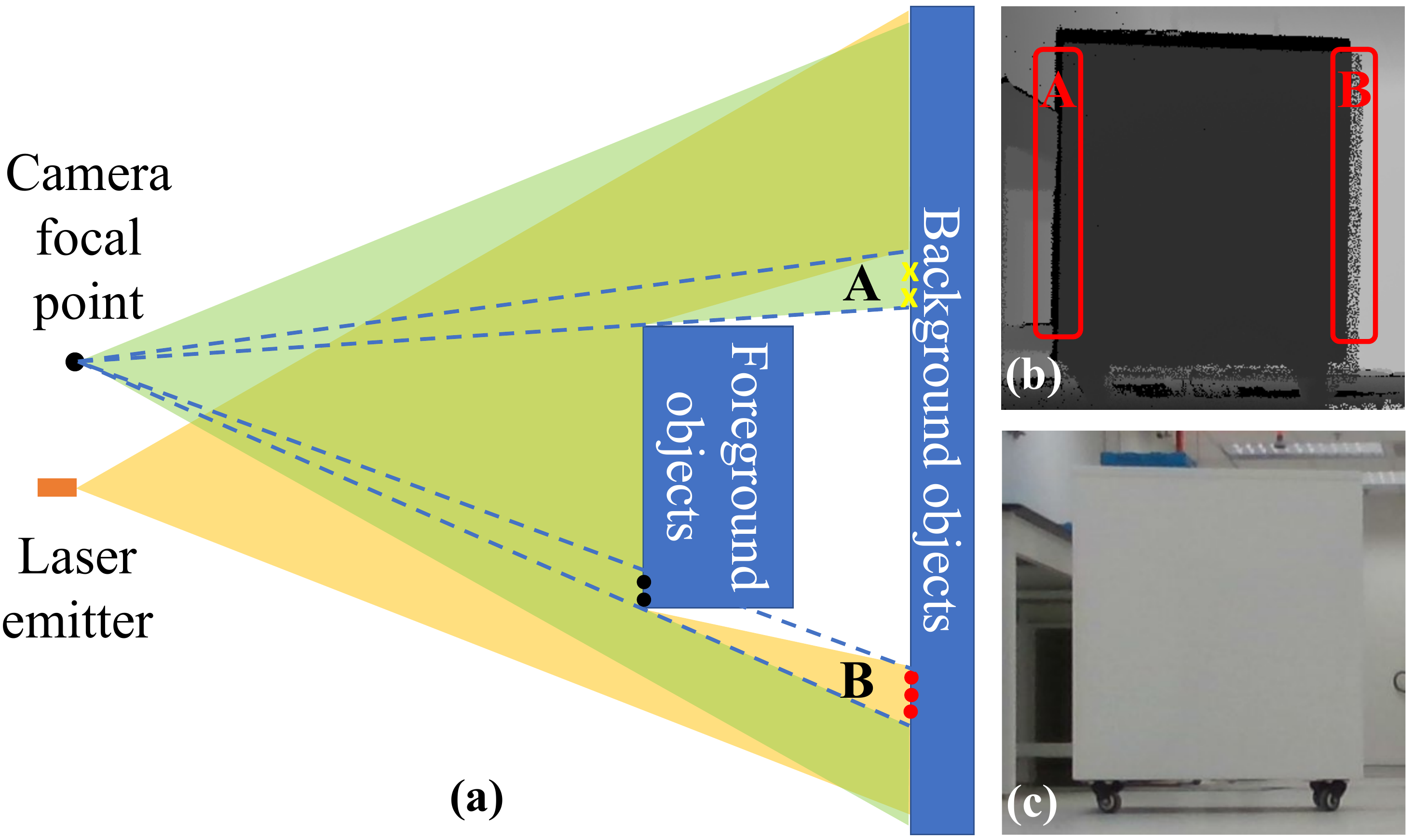}
    \caption{Multi-valued mapping and zero-valued mapping.}
    \label{fig:view_error}
\end{figure}

Some existing works project the point cloud to the image plane and extract features from the projected point cloud, such as edge extraction \cite{zhu2020camvox} and the mutual information correlation \cite{pandey2012automatic}. A major problem of the feature extraction after points projection is the multi-valued and zero-valued mapping caused by occlusion. As illustrated in Fig. \ref{fig:view_error} (a), if the camera is above the LiDAR, region A is observed by the camera but not LiDAR due to occlusion, resulting in no points after projection in this region (zero-valued mapping, the gap in region A, Fig. \ref{fig:view_error} (b)). On the other hand, region B is observed by the LiDAR but not the camera, points in this region (the red dots in Fig. \ref{fig:view_error} (a)) after projection will intervene the projection of points on its foreground (the black dots in Fig. \ref{fig:view_error} (a)). As a result, points of the foreground and background will correspond to the same image regions (multi-valued mapping, region B, Fig. \ref{fig:view_error} (b)). These phenomenon may not be significant for LiDARs of low-resolution \cite{pandey2012automatic}, but is evident in as the LiDAR resolution increases (see Fig. \ref{fig:view_error} (b)).  Extracting features on the projected point clouds and match them to the image features, such as \cite{pandey2012automatic},  would suffer from these fundamental problems and cause significant errors in edge extraction and calibration. 

\begin{figure}[t]
    \centering
    \includegraphics[width=0.6\linewidth]{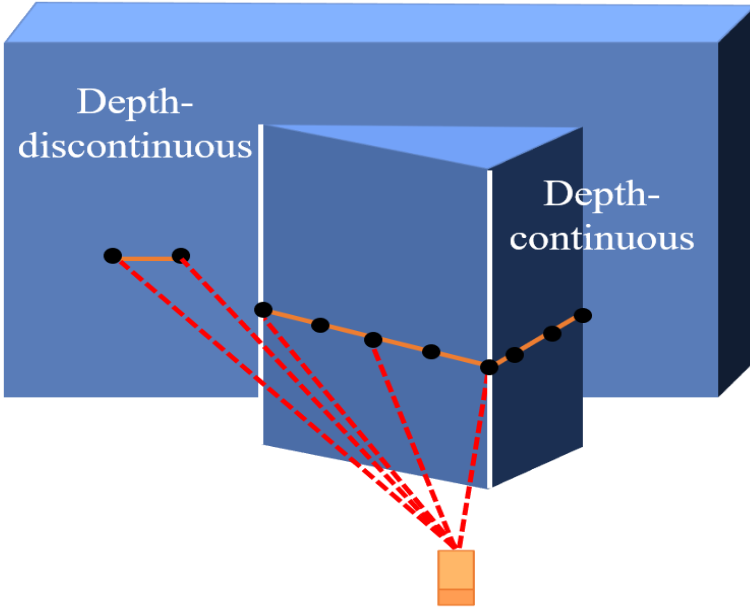}
    \caption{Depth-discontinuous edge and depth-continuous edge.}
    \label{fig:edge_type}
    \vspace{-0.3cm}
\end{figure}

\begin{figure}[t]
    \vspace{-0.3cm}
    \centering
    \includegraphics[width=1\linewidth]{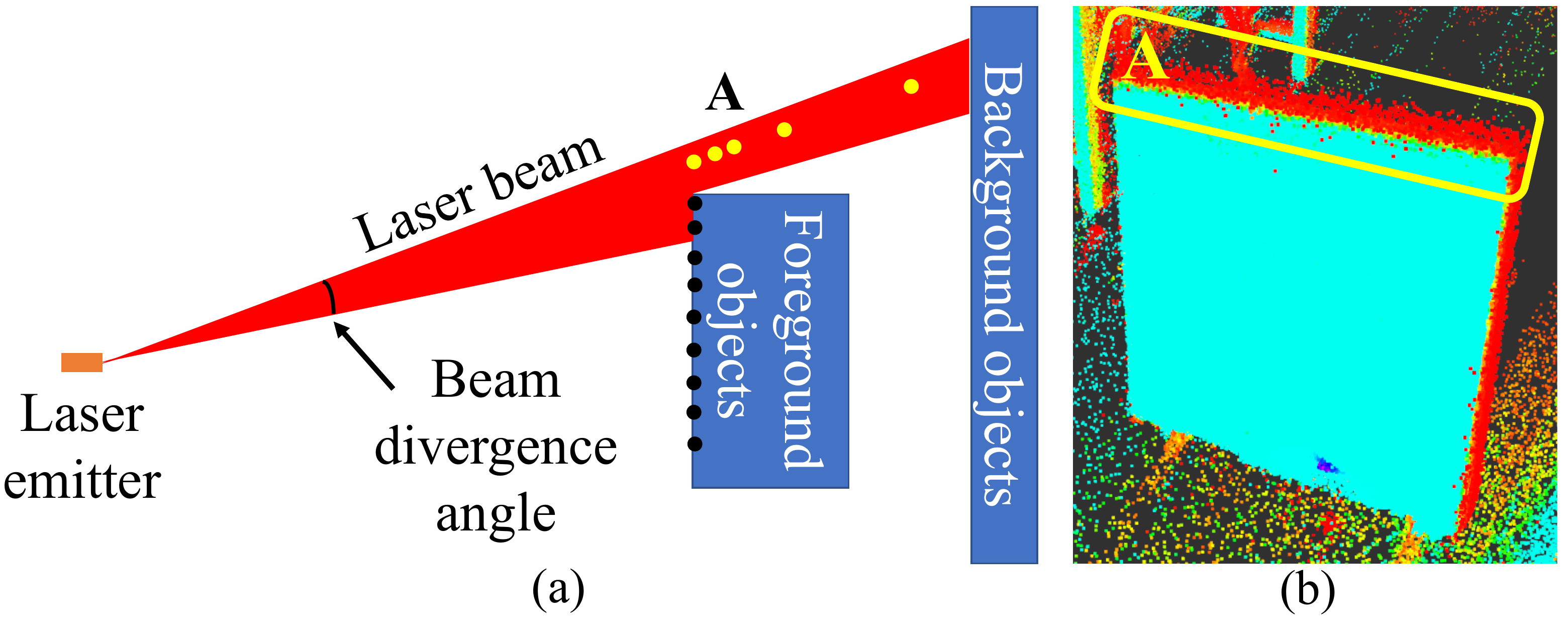}
    \caption{Foreground objects inflation and bleeding points caused by laser beam divergence angle.}
    \label{fig:bleeding_points}
\end{figure}

To avoid the zero-valued and multi-valued mapping problem caused by projection, we extract edge features directly on the LiDAR point cloud. There are two types of edges: depth-discontinuous and depth-continuous. As shown in Fig. \ref{fig:edge_type} (a), depth-discontinuous edges refer to edges between foreground objects and background objects where the depth jumps. In contrast, depth-continuous edges refer to the plane joining lines where the depth varies continuously. Many existing methods \cite{scaramuzza2007extrinsic, levinson2013automatic} use the depth-discontinuous edges as they can be easily extracted by examining the point depth. However, carefully investigating the LiDAR measuring principle, we found that depth-discontinuous edges are not reliable nor accurate for high accuracy calibration. As shown in Fig. \ref{fig:bleeding_points}, a practical laser pulse is not an ideal point but a beam with a certain divergence angle (i.e., the beam divergence angle). When scanning from a foreground object to a background one, part of the laser pulse is reflected by the foreground object while the remaining reflected by the background, producing two reflected pulses to the laser receiver. In the case of the high reflectivity of the foreground object, signals caused by the first pulse will dominate, even when the beam centerline is off the foreground object, this will cause fake points of the foreground object beyond the actual edge (the leftmost yellow point in Fig. \ref{fig:bleeding_points} (a)). In the case the foreground object is close to the background, signals caused by the two pulses will join, and the lumped signal will lead to a set of points connecting the foreground and the background (called the {\it bleeding points}, the yellow points in Fig. \ref{fig:bleeding_points} (a)). The two phenomena will mistakenly inflate the foreground object and cause significant errors in the edge extraction (Fig. \ref{fig:bleeding_points} (b)) and calibration. 

\begin{figure}
    \centering
    \includegraphics[width=1\linewidth]{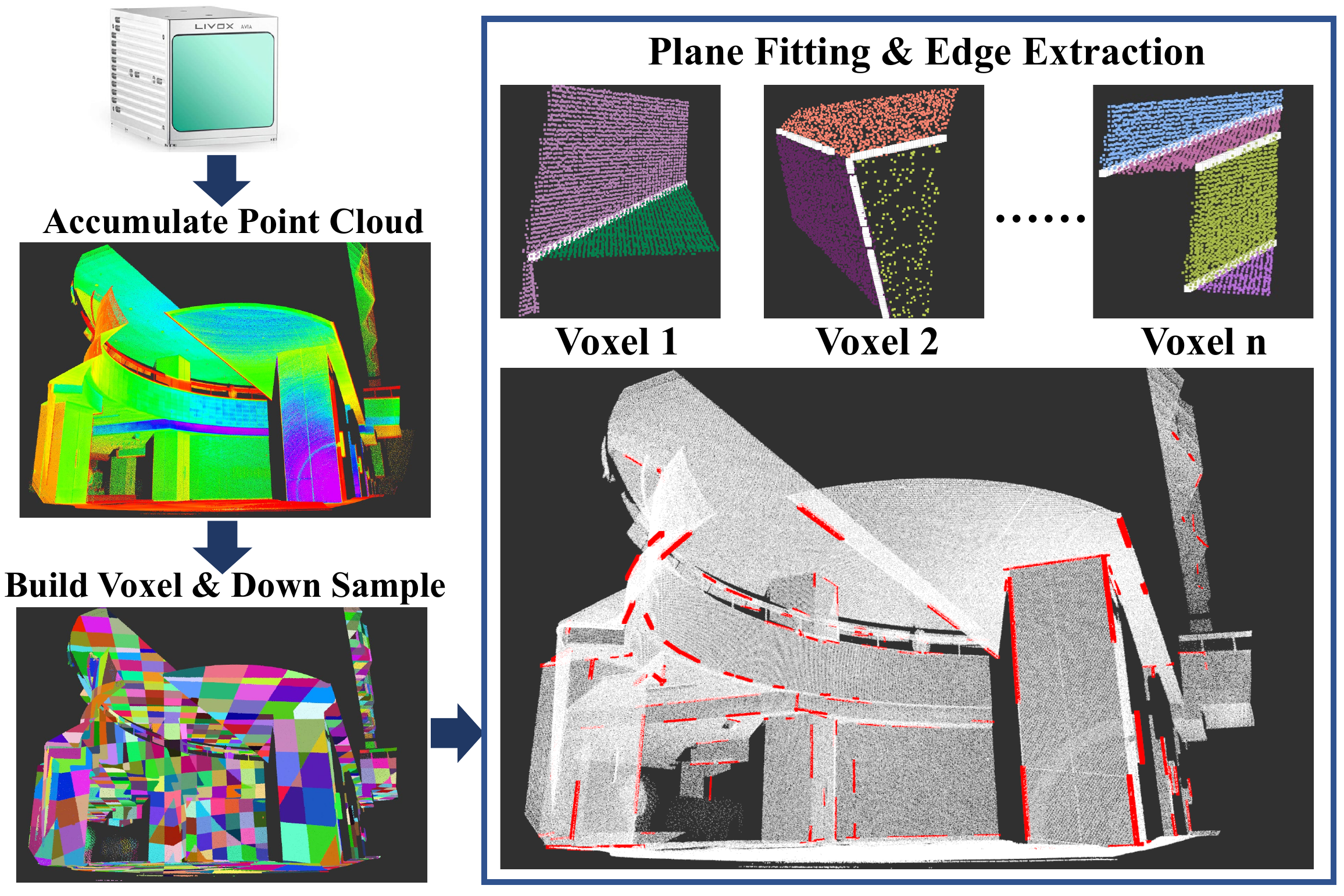}
    \caption{Depth-continuous edge extraction. In each voxel grid, different colors represent different voxels. Within the voxel, different colors represent different planes, and the white lines represent the intersections between planes.}
    \label{fig:extract_edge}
    \vspace{-0.3cm}
\end{figure}

To avoid the foreground inflation and bleeding points caused by depth-discontinuous edges, we propose to extracting depth-continuous edges. The overall procedure is summarized in Fig.~\ref{fig:extract_edge}: we first divide the point cloud into small voxels of given sizes (e.g., $1m$ for outdoor scenes and $0.5m$ for indoor scenes). For each voxel, we repeatedly use RANSAC to fit and extract planes contained in the voxel. Then, we retain plane pairs that are connected and form an angle within a certain range (e.g., $[\ang{30},\ang{150}]$) and solve for the plane intersection lines (i.e., the depth-continuous edge). As shown in Fig. \ref{fig:extract_edge}, our method is able to extract multiple intersection lines that are perpendicular or parallel to each other within a voxel. Moreover, by properly choosing the voxel size, we can even extract curved edges.

\begin{figure}
    \centering
    \vspace{-0.2cm}
    \includegraphics[width=1\linewidth]{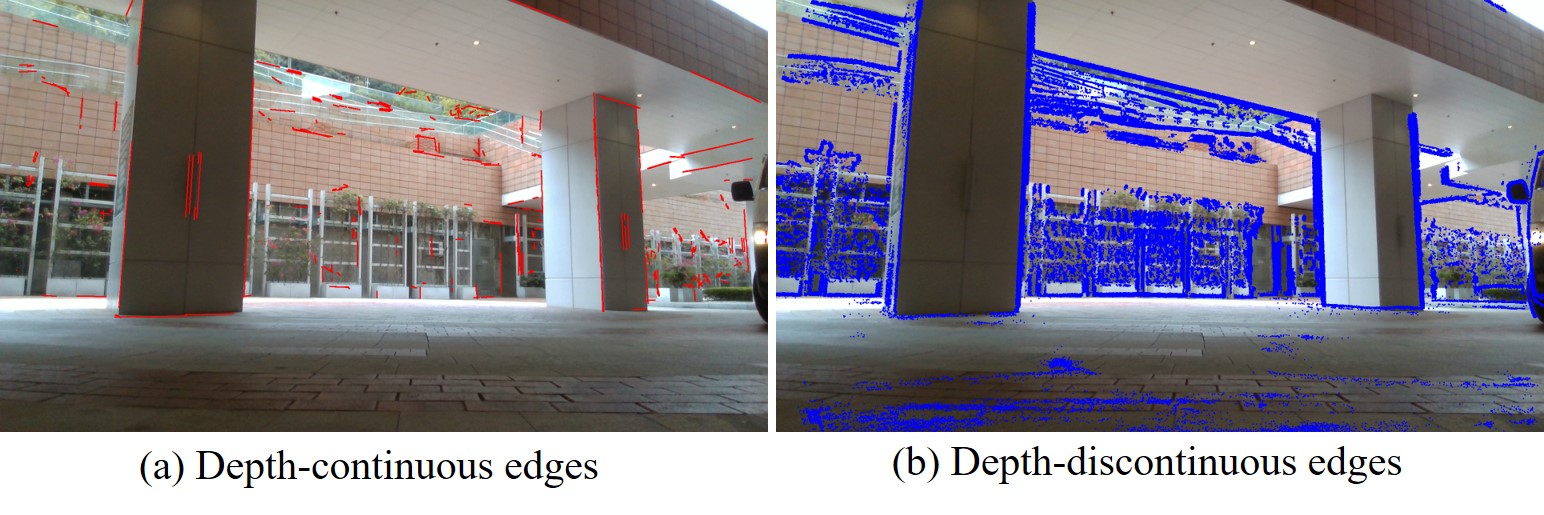}
    \caption{\textcolor{black}{Comparison between extracted depth-continuous edges and depth-discontinuous edges.}}
    \label{fig:edge_contrast}
    \vspace{-0.3cm}
\end{figure}

\textcolor{black}{Fig. \ref{fig:edge_contrast} shows a comparison between the extracted depth-continuous and depth-discontinuous edges when overlaid with the image using correct extrinsic value. The depth-discontinuous edge is extracted based on the local curvature as in \cite{liu2021balm}. It can be seen that the depth-continuous edges are more accurate and contain less noise.
}

For image edge extraction, we use the Canny algorithm~\cite{canny}. The extracted edge pixels are saved in a $k$-D tree ($k=2$) for the correspondence matching.

\subsubsection{Matching}\label{sec:matching}

The extracted LiDAR edges need to be matched to their corresponding edges in the image. For each extracted LiDAR edge, we sample multiple points on the edge. Each sampled point ${}^L \mathbf P_i \in \mathbb{R}^3$ is transformed into the camera frame (using the current extrinsic estimate ${}^C_L \bar{\mathbf T} = ({}^C_L \bar{\mathbf R} , {}^C_L \bar{\mathbf t} ) \in SE(3)$)
\vspace{-0.1cm}
\begin{equation}
    {}^C \mathbf P_i = {}^C_L \bar{\mathbf T} ( {}^L \mathbf P_i ) \in \mathbb{R}^3,
    \vspace{-0.2cm}
    \label{eq:transform}
\end{equation}
where ${}_L^C \bar{\mathbf T}({}^L \mathbf P_i ) = {}_L^C \bar{\mathbf R} \cdot {}^L\mathbf P_i + {}_L^C \bar{\mathbf t}$ denotes applying the rigid transformation ${}_L^C \mathbf T $ to point ${}^L \mathbf P_i$. The transformed point ${}^C \mathbf P_i$ is then projected onto the camera image plane to produce an expected location ${}^C \mathbf p_i \in \mathbb{R}^2$
\vspace{-0.2cm}
\begin{equation}
    {}^C \mathbf p_i = \boldsymbol{\pi}({}^C \mathbf P_i)
    \vspace{-0.1cm}
\end{equation}
where $\boldsymbol{\pi}(\mathbf P)$ is the pin-hole projection model. Since the actual camera is subject to distortions, the actual location of the projected point $\mathbf{p}_i = ( u_i, v_i)$ on the image plane is
\vspace{-0.1cm}
\begin{equation}
    \mathbf p_i = \mathbf f ({}^C \mathbf p_i)
    \label{eq:distort}
    \vspace{-0.1cm}
\end{equation}
where $\mathbf f(\mathbf p)$ is the camera distortion model. 

We search the $\kappa$ nearest neighbor of $\mathbf p_{i}$ in the $k$-D tree built from the image edge pixels. Denotes $\mathbf Q_i=\{\mathbf q_i^j; j = 1, \cdots, \kappa\}$ the $\kappa$ nearest neighbours and
\begin{equation}
\begin{aligned}
    \mathbf q_i = \frac{1}{\kappa} \sum_{j=1}^\kappa \mathbf q_{i}^j; \quad \mathbf{S}_i = \sum_{j=1}^{\kappa} (\mathbf {q}_i^j - \mathbf q_i ) (\mathbf {q}_i^j - \mathbf q_i )^T.
\end{aligned}
\end{equation}
Then, the line formed by $\mathbf Q_i$ is parameterized by point $\mathbf q_i$ lying on the line and the normal vector $\mathbf n_i$, which is the eigenvector associated to the minimal eigenvalue of $\mathbf S_i$. 

\begin{figure}[t]
    \vspace{-0.3cm}
    \centering
    \includegraphics[width=1\linewidth]{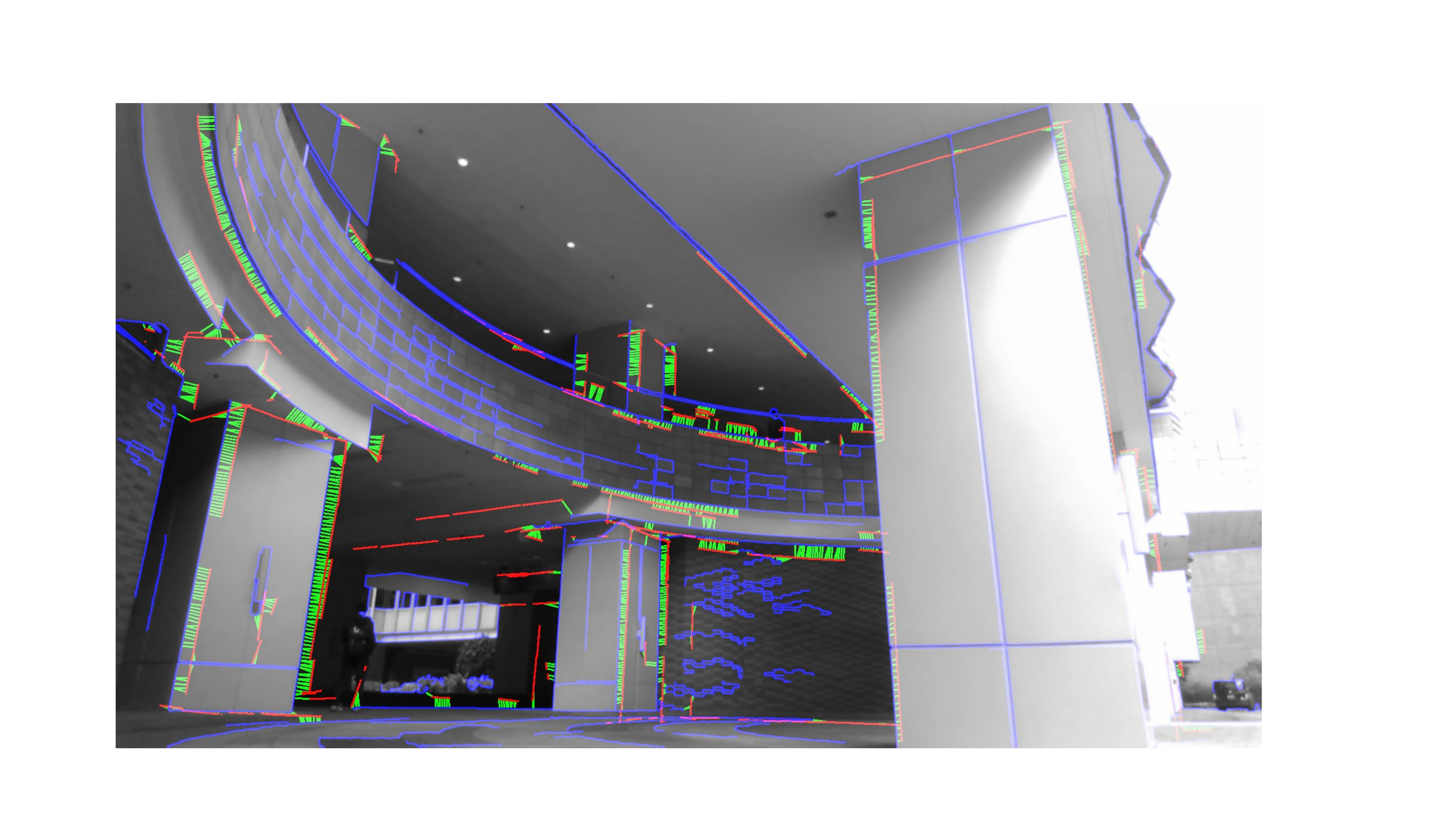}
    \caption{LiDAR edges (red lines), image edge pixels (blue lines) and their correspondences (green lines).}
    \label{fig:lidar_camera_edge}
    \vspace{-0.5cm}
\end{figure}

Besides projecting the point ${}^L \mathbf P_i$ sampled on the extracted LiDAR edge, we also project the edge direction to the image plane and validate its orthogonality with respect to $\mathbf n_i$. This can effectively remove \textcolor{black}{error matches} when two non-parallel lines are near to each other in the image plane. Fig. \ref{fig:lidar_camera_edge} shows an example of the extracted LiDAR edges (red lines), image edge pixels (blue lines) and the correspondences (green lines).

\vspace{-0.2cm}\subsection{Extrinsic Calibration}
\subsubsection{Measurement noises}

The extracted LiDAR edge points ${}^L \mathbf P_i$ and the corresponding edge feature $(\mathbf n_i, \mathbf q_i)$ in the image are subject to measurement noises. Let ${}^I\mathbf w_{i} \in \mathcal{N}(\mathbf 0, {}^I \boldsymbol{\Sigma}_i) $ be the noise associated with $\mathbf q_i$ during the image edge extraction, its covariance is ${}^I \boldsymbol{\Sigma}_i = \sigma_I^2 \mathbf I_{2 \times 2}$, where $\sigma_I = 1.5$ indicating the one-pixel noise due to pixel discretization. 

\begin{figure}[t]
    \vspace{-0.3cm}
    \centering
    \includegraphics[width=1\linewidth]{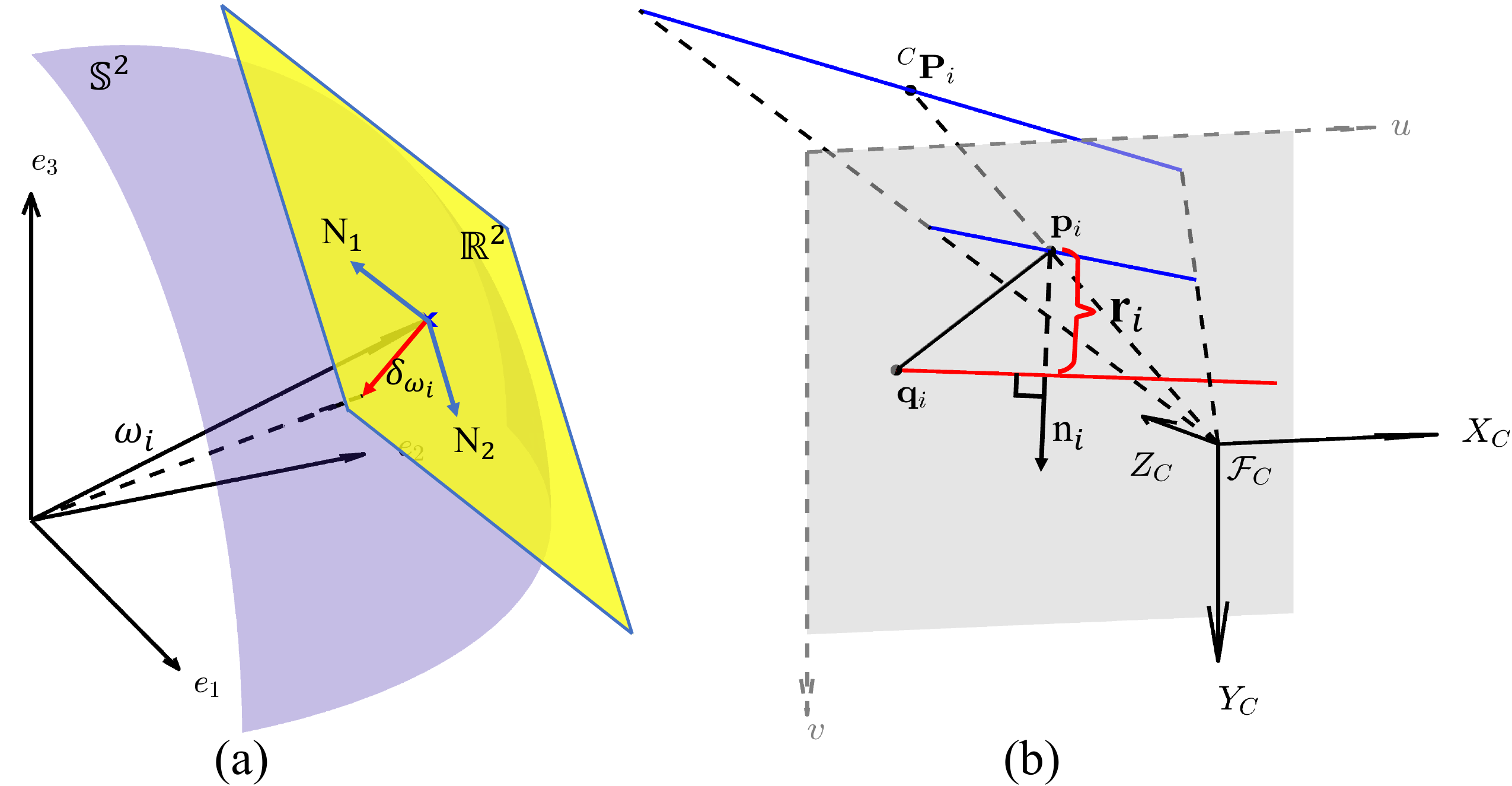}
    \caption{(a) Perturbation to a bearing vector on $\mathbb{S}^2$; (b) Projection of a LiDAR edge point, expressed in the camera frame ${}^C \mathbf P_i$, to the image plane $\mathbf p_i$, and the computation of residual $\mathbf r_i$.}
    \label{fig:perturbation}
    \vspace{-0.6cm}
\end{figure}

For the LiDAR point ${}^L \mathbf P_i$, let ${}^L\mathbf w_{i}$ be its measurement noise. In practice, LiDAR measures the bearing direction by encoders of the scanning motor and the depth by computing the laser time of flight. Let $\boldsymbol{\omega}_i \in \mathbb{S}^2$ be the measured bearing direction and $\boldsymbol{\delta}_{\boldsymbol{\omega}_i} \sim \mathcal{N}(\mathbf 0_{2\times 1}, \boldsymbol{\Sigma}_{\boldsymbol{\omega}_i})$ be the measurement noise in the tangent plane of $\boldsymbol{\omega}_i$ (see Fig. \ref{fig:perturbation}). Then using the $\boxplus$-operation encapsulated in $\mathbb{S}^2$ \cite{he2021embedding}, we obtain the relation between the true bearing direction $\boldsymbol{\omega}_i^{\text{gt}}$ and its measurement $\boldsymbol{\omega}_i$ as below:
\begin{equation}
    \label{eq:bearing_model}
    \boldsymbol{\omega}_i^{\text{gt}} = \boldsymbol{\omega}_i \boxplus_{\mathbb{S}^2} \boldsymbol{\delta}_{\boldsymbol{\omega}_i}  \triangleq e^{ \lfloor \mathbf N(\boldsymbol{\omega}_i) \boldsymbol{\delta}_{\boldsymbol{\omega}_i} \times \rfloor} \boldsymbol{\omega}_i
\end{equation}
where $\mathbf N(\boldsymbol{\omega}_i) = \begin{bmatrix}
\mathbf N_1 & \mathbf N_2
\end{bmatrix} \in \mathbb{R}^{3 \times 2} $ is an orthonormal basis of the tangent plane at $\boldsymbol{\omega}_i$ (see Fig. \ref{fig:perturbation} (a)), and $\lfloor  \  \times \rfloor$ denotes the skew-symmetric matrix mapping the cross product. The $\boxplus_{\mathbb{S}^2}$-operation essentially rotates the unit vector $\boldsymbol{\omega}_i$ about the axis $ \boldsymbol{\delta}_{\boldsymbol{\omega}_i}$ in the tangent plane at $\boldsymbol{\omega}_i$, the result is still a unit vector (i.e., remain on $\mathbb{S}^2$). 

Similarly, let $d_i$ be the depth measurement and $\delta_{d_i} \sim \mathcal{N}(0, {\Sigma}_{{d}_i}) $ be the ranging error, then the ground-\textcolor{black}{truth} depth $d_i^{\text{gt}}$ is 
\begin{equation}
    \label{eq:ranging_model}
    d_i^{\text{gt}} = d_i + \delta_{d_i}
\end{equation}

Combining (\ref{eq:bearing_model}) and (\ref{eq:ranging_model}), we obtain the relation between the ground-\textcolor{black}{truth} point location ${}^L \mathbf P_i^{\text{gt}}$ and its measurement ${}^L \mathbf P_i$:
\begin{equation}
    \begin{split}
        {}^L \mathbf P_i^{\text{gt}} &= d_i^{\text{gt}} \boldsymbol{\omega}_i^{\text{gt}} = \left( d_i + \delta_{d_i} \right) \left( \boldsymbol{\omega}_i \boxplus_{\mathbb{S}^2} \boldsymbol{\delta}_{\boldsymbol{\omega}_i} \right) \\
        &\approx \underbrace{d_i \boldsymbol{\omega}_i}_{{}^L \mathbf P_i} + \underbrace{ \boldsymbol{\omega}_i \delta_{d_i} - d_i \lfloor \boldsymbol{\omega}_i \times \rfloor \mathbf N(\boldsymbol{\omega}_i) \boldsymbol{\delta}_{\boldsymbol{\omega}_i}}_{{}^L \mathbf w_i} 
    \end{split}
\end{equation}
Therefore,
\begin{equation}
    \begin{split}
        {}^L \mathbf w_i &= \underbrace{\begin{bmatrix}
        \boldsymbol{\omega}_i & - d_i \lfloor \boldsymbol{\omega}_i \times \rfloor \mathbf N(\boldsymbol{\omega}_i)
    \end{bmatrix}}_{\mathbf A_i }
    \begin{bmatrix}
     \delta_{d_i} \\
     \boldsymbol{\delta}_{\boldsymbol{\omega}_i}
    \end{bmatrix} \sim \mathcal{N}(\mathbf 0, {}^L \boldsymbol{\Sigma}_i), \\
    {}^L \boldsymbol{\Sigma}_i &= \mathbf A_i 
    \begin{bmatrix}
    \Sigma_{d_i} & \mathbf 0_{1 \times 2}  \\
    \mathbf 0_{2 \times 1} & \boldsymbol{\Sigma}_{\boldsymbol{\omega}_i}
    \end{bmatrix}\mathbf A_i^T.
    \end{split}
    \label{eq:meas_noise_cov}
\end{equation}

This noise model will be used to produce a consistent extrinsic calibration as detailed follow.

\subsubsection{Calibration Formulation and Optimization}\label{sec:formulation}

Let ${}^L \mathbf P_{i}$ be an edge point extracted from the LiDAR point cloud and its corresponding edge in the image is represented by its normal vector $\mathbf n_i \in \mathbb{S}^1 $ and a point $\mathbf q_i \in \mathbb{R}^2$ lying on the edge (Section \ref{sec:matching}). Compensating the noise in ${}^L \mathbf P_{i}$  and projecting it to the image plane using the \textcolor{black}{ground-truth} extrinsic should \textcolor{black}{lie exactly on the edge} $(\mathbf n_i, \mathbf q_i)$ extracted from the image (see (\ref{eq:transform} -- \ref{eq:distort}) and Fig. \ref{fig:perturbation} (b)): 
\begin{equation}
0 =  \mathbf n_i^T \left( \mathbf f \!  \left(\boldsymbol{\pi} \! \left({}^C_L \mathbf T \left( {}^L\mathbf P_{i} \! + \! {}^L\mathbf w_{i} \right) \right) \right) \! - \!  \left( \mathbf q_i \! + \! {}^I \mathbf w_i \right) \right)
\label{eq:formulation}
\end{equation}
where ${}^L\mathbf w_{i} \in \mathcal{N}(\mathbf 0, {}^L \boldsymbol{\Sigma}_i) $ and ${}^I\mathbf w_{i} \in \mathcal{N}(\mathbf 0, {}^I \boldsymbol{\Sigma}_i) $ are detailed in the previous section. 

Equation (\ref{eq:formulation}) implies that one LiDAR edge point imposes one constraint to the extrinsic, which is in agreement with Section \ref{sec:overview} that an edge feature imposes two constraints to the extrinsic as an edge consists of two independent points. Moreover, (\ref{eq:formulation}) imposes a nonlinear equation for the extrinsic ${}^C_L \mathbf T$ in terms of the measurements ${}^L \mathbf P_{i}, \mathbf n_i, \mathbf q_i$ and unknown noise ${}^L\mathbf w_{i}, {}^I\mathbf w_{i}$. This nonlinear equation can be solved in an iterative way: let ${}^C_L \bar{\mathbf T}$ be the current extrinsic estimate and parameterize ${}^C_L \mathbf T$ in the tangent space of  ${}^C_L \bar{\mathbf T}$ using the $\boxplus$-operation encapsulated in $SE(3)$ \cite{hertzberg2013integrating, he2021embedding}:
\begin{equation}
    {}^C_L \mathbf T = {}^C_L \bar{\mathbf T} \boxplus_{SE(3)} \delta \mathbf T \triangleq \text{Exp}(\delta \mathbf T) \cdot {}^C_L \bar{\mathbf T}
    \label{e:local_param}
\end{equation}
where 
$$ \delta \mathbf T = \begin{bmatrix}
\delta \boldsymbol{\theta} \\
\delta \mathbf t 
\end{bmatrix} \in \mathbb{R}^6; \ \text{Exp}(\delta \mathbf T) = 
\begin{bmatrix}
e^{\lfloor \delta \boldsymbol{\theta} \times \rfloor} & \delta \mathbf t \\
0 & 1
\end{bmatrix} \in SE(3). $$

Substituting (\ref{e:local_param}) into (\ref{eq:formulation}) and approximating the resultant equation with first order terms lead to
\begin{equation}
    \begin{split}
         0  &= \mathbf n_i^T  \left(\mathbf f \!  \left(\boldsymbol{\pi} \! \left({}^C_L \mathbf T \left( {}^L\mathbf P_{i} \! + \! {}^L\mathbf w_{i} \right) \right) \right) \! - \!  \left( \mathbf q_i \! + \! {}^I \mathbf w_i \right) \right) \\
    &\approx \mathbf r_i + \mathbf J_{\mathbf T_i} \delta \mathbf T + \mathbf J_{\mathbf w_i} \mathbf w_i
    \end{split}
    \label{eq:approx}
\end{equation}
where 
\begin{equation}
    \begin{split}
        \mathbf r_i & = \mathbf n_i^T \left( \mathbf f \!  \left(\boldsymbol{\pi} \! \left({}^C_L \bar{\mathbf T}  ({}^L\mathbf P_{i})  \right) \right) \! - \!   \mathbf q_i \right) \in \mathbb{R} \\
        \mathbf J_{\mathbf T_i} &= \mathbf n_i^T \frac{\partial \mathbf f (\mathbf p)}{\partial \mathbf p} \frac{\partial \boldsymbol{\pi} (\mathbf P)}{\partial \mathbf P} 
\begin{bmatrix}
-\lfloor ({}^C_L \bar{\mathbf T}  ({}^L \mathbf P_i)) \times \rfloor & \mathbf I
\end{bmatrix}\in \mathbb{R}^{1 \times 6} \\
        \mathbf J_{\mathbf w_i} &= \begin{bmatrix}
        \mathbf n_i^T \frac{\partial \mathbf f (\mathbf p)}{\partial \mathbf p} \frac{\partial \boldsymbol{\pi} (\mathbf P)}{\partial \mathbf P} {}^C_L \bar{\mathbf R} & -\mathbf n_i^T
        \end{bmatrix} \in \mathbb{R}^{1 \times 5} \\
        \mathbf w_i &= \begin{bmatrix}
        {}^L \mathbf w_i \\
        {}^I \mathbf w_i
        \end{bmatrix} \in \mathcal{N}(\mathbf 0, \boldsymbol{\Sigma}_i), \boldsymbol{\Sigma}_i = \begin{bmatrix}
        {}^L \boldsymbol{\Sigma}_i & \mathbf 0 \\
        \mathbf 0 & {}^I \boldsymbol{\Sigma}_i
        \end{bmatrix} \in \mathbb{R}^{5 \times 5}
    \end{split}
    \label{eq:grad}
\end{equation}

The calculation of $\mathbf r_i$ is illustrated in Fig. \ref{fig:perturbation} (b). Equation (\ref{eq:approx}) defines the constraint from one edge correspondence, stacking all $N$ such edge correspondences leads to
\begin{equation}
    \begin{split}
         \underbrace{\begin{bmatrix}
         0 \\
         \vdots \\
         0
         \end{bmatrix}}_{\mathbf 0}  &\approx \underbrace{\begin{bmatrix}
         \mathbf r_1 \\
         \vdots \\
         \mathbf r_N
         \end{bmatrix}}_{\mathbf r} + \underbrace{
\begin{bmatrix}
\mathbf J_{\mathbf T_1} \\
\vdots \\
\mathbf J_{\mathbf T_N}
\end{bmatrix}}_{\mathbf J_{\mathbf T}} \delta \mathbf T +  \underbrace{ \begin{bmatrix}
\mathbf J_{\mathbf w_1} & \cdots & \mathbf 0 \\
\vdots & \ddots & \vdots \\
\mathbf 0 & \cdots & \mathbf J_{\mathbf w_N}
\end{bmatrix}}_{\mathbf J_{\mathbf w}} \underbrace{\begin{bmatrix}
\mathbf w_1 \\
\vdots \\
\mathbf w_N
\end{bmatrix}}_{\mathbf w}
    \end{split}
    \label{eq:all_eqn}
\end{equation}
where$$\mathbf w \sim \mathcal{N}(\mathbf 0, \boldsymbol{\Sigma}), \ \boldsymbol{\Sigma} = \text{diag}(\boldsymbol{\Sigma}_1, \cdots, \boldsymbol{\Sigma}_N)$$

Equation (\ref{eq:all_eqn}) implies
\begin{equation}
    \begin{split}
        \mathbf v \triangleq - \mathbf J_{\mathbf w} \mathbf w = \mathbf r + \mathbf J_{\mathbf T} \delta \mathbf T \sim \mathcal{N}(\mathbf 0, \mathbf J_{\mathbf w} \boldsymbol{\Sigma} \mathbf J_{\mathbf w}^T).
        \label{eq:distr}
    \end{split}
\end{equation}
Based on (\ref{eq:distr}), we propose our maximal likelihood (and meanwhile the minimum variance) extrinsic estimation:
\begin{equation}
    \begin{split}
        & \max_{\delta \mathbf T} \log p(\mathbf v; \delta \mathbf T) = \max_{\delta \mathbf T} \log \frac{e^{-\frac{1}{2} \mathbf v^T \left( \mathbf J_{\mathbf w} \boldsymbol{\Sigma} \mathbf J_{\mathbf w}^T \right)^{-1} \mathbf v }}{\sqrt{(2 \pi)^N \det \left( \mathbf J_{\mathbf w} \boldsymbol{\Sigma} \mathbf J_{\mathbf w}^T \right)}}  \\
        &= \min_{\delta \mathbf T} (\mathbf r + \mathbf J_{\mathbf T} \delta \mathbf T)^T \left( \mathbf J_{\mathbf w} \boldsymbol{\Sigma} \mathbf J_{\mathbf w}^T \right)^{-1} (\mathbf r + \mathbf J_{\mathbf T} \delta \mathbf T)
    \end{split}
    \label{eq:cost_func}
\end{equation}

The optimal solution is 
\begin{equation}
    \delta \mathbf T^* = - \left( \mathbf J_{\mathbf T}^T \left( \mathbf J_{\mathbf w} \boldsymbol{\Sigma} \mathbf J_{\mathbf w}^T \right)^{-1} \mathbf J_{\mathbf T} \right)^{-1} \mathbf J_{\mathbf T}^T \left( \mathbf J_{\mathbf w} \boldsymbol{\Sigma} \mathbf J_{\mathbf w}^T \right)^{-1} \mathbf r
    \label{eq:Tstar}
\end{equation}

This solution is updated to ${}^C_L \bar{\mathbf T}$ 
\begin{equation}
    {}^C_L \bar{\mathbf T} \leftarrow {}^C_L \bar{\mathbf T} \boxplus_{SE(3)} \delta \mathbf T^*.
    \label{eq:update}
\end{equation}
The above process ((\ref{eq:Tstar} and (\ref{eq:update})) iterates until convergence (i.e., $\| \delta \mathbf T^* \| < \varepsilon$) and the converged ${}^C_L \bar{\mathbf T}$ is the calibrated extrinsic.

\subsubsection{Calibration Uncertainty}

Besides the extrinsic calibration, it is also useful to estimate the calibration uncertainty, which can be characterized by the covariance of the error between the ground-\textcolor{black}{truth} extrinsic and the calibrated one. To do so, we multiply both sides of (\ref{eq:all_eqn}) by $ \mathbf J_{\mathbf T}^T \left( \mathbf J_{\mathbf w} \boldsymbol{\Sigma} \mathbf J_{\mathbf w}^T \right)^{-1} $ and solve for $\delta \mathbf T$:
\begin{equation}
    \begin{split}
        \delta \mathbf T &\approx \underbrace{-\left( \mathbf J_{\mathbf T}^T \left( \mathbf J_{\mathbf w} \boldsymbol{\Sigma} \mathbf J_{\mathbf w}^T \right)^{-1} \mathbf J_{\mathbf T} \right)^{-1} \mathbf J_{\mathbf T}^T \left( \mathbf J_{\mathbf w} \boldsymbol{\Sigma} \mathbf J_{\mathbf w}^T \right)^{-1} \mathbf r}_{\delta \mathbf T^*} \\
        &-  \left( \mathbf J_{\mathbf T}^T \left( \mathbf J_{\mathbf w} \boldsymbol{\Sigma} \mathbf J_{\mathbf w}^T \right)^{-1} \mathbf J_{\mathbf T} \right)^{-1}  \mathbf J_{\mathbf T}^T \left( \mathbf J_{\mathbf w} \boldsymbol{\Sigma} \mathbf J_{\mathbf w}^T \right)^{-1} \mathbf J_{\mathbf w} \mathbf w \\
        &\sim \mathcal{N} \left(\delta \mathbf T^*, \left( \mathbf J_{\mathbf T}^T \left( \mathbf J_{\mathbf w} \boldsymbol{\Sigma} \mathbf J_{\mathbf w}^T \right)^{-1} \mathbf J_{\mathbf T} \right)^{-1} \right)
    \end{split}
    \label{eq:Cov}
\end{equation}
which means that the ground truth $\delta \mathbf T$, the error between the ground truth extrinsic ${}^C_L \mathbf T$ and the estimated one ${}^C_L \bar{\mathbf T}$ and parameterized in the tangent space of ${}^C_L \bar{\mathbf T}$,  is subject to a Gaussian distribution that has a mean $\delta \mathbf T^*$ and covariance equal to the inverse of the Hessian matrix of (\ref{eq:cost_func}). At convergence, the  $\delta \mathbf T^*$ is near to zero, and the covariance is
\begin{equation}
    \boldsymbol{\Sigma}_{\mathbf T} = \left( \mathbf J_{\mathbf T}^T \left( \mathbf J_{\mathbf w} \boldsymbol{\Sigma} \mathbf J_{\mathbf w}^T \right)^{-1} \mathbf J_{\mathbf T} \right)^{-1}
    \label{eq:estiamte_cov}
\end{equation}
We use this covariance matrix to characterize  our extrinsic calibration uncertainty.

\vspace{-0.3cm}\subsection{Analysis of Edge Distribution on Calibration Result}\label{sec:analysis}

The Jacobian $\mathbf J_{\mathbf T_i}$ in (\ref{eq:grad}) denotes the sensitivity of residual with respect to the extrinsic variation. In case of very few or poorly distributed edge features, $\mathbf J_{\mathbf T_i}$ could be very small, leading to large estimation uncertainty (covariance) as shown by (\ref{eq:estiamte_cov}). In this sense, the data quality is automatically \textcolor{black}{and quantitatively} encoded by the covariance matrix in (\ref{eq:estiamte_cov}). In practice, it is usually useful to have a quick and rough assessment of the calibration scene before the data collection. This can be achieved by analytically deriving the Jacobian $\mathbf J_{\mathbf T_i}$. Ignoring the distortion model and substituting the pin-hole projection model $\boldsymbol{\pi}(\cdot)$ to (\ref{eq:grad}), we obtain
\begin{equation}
\begin{aligned}
    \mathbf J_{T_{i}} 
    &= \mathbf n_{i}^T
    \begin{bmatrix}
    \frac{-f_x X_{i}{ Y_i}}{ Z_i^2}\hspace{-0.2cm}&\hspace{-0.2cm}f_x+\frac{f_x  X_i^2}{ Z_i^2}\hspace{-0.2cm}&\hspace{-0.2cm}\frac{-f_x{ Y_i}}{{Z_i}}\hspace{-0.2cm}&\hspace{-0.2cm}\frac{f_x}{{ Z_i}}\hspace{-0.2cm}&\hspace{-0.2cm}0\hspace{-0.2cm}&\hspace{-0.2cm}-\frac{f_x { X_i}}{ Z_i^2}\\
    -f_y-\frac{f_y Y_i^2}{Z_i^2}\hspace{-0.2cm}&\hspace{-0.2cm}\frac{f_y {X_i} {Y_i}}{Z_i^2}\hspace{-0.2cm}&\hspace{-0.2cm}\frac{f_y { X_{i}}}{{ Z_i}}\hspace{-0.2cm}&\hspace{-0.2cm}0\hspace{-0.2cm}&\hspace{-0.2cm}\frac{f_y}{{ Z_i}}\hspace{-0.2cm}&\hspace{-0.2cm}\frac{-f_y { Y_i}}{ Z_i^2}
    \end{bmatrix}
\end{aligned}
\end{equation}
where ${}^C \mathbf P_i = \begin{bmatrix}
X_i & Y_i & Z_i
\end{bmatrix}^T$ is the LiDAR edge point ${}^L \mathbf P_i$ represented in the camera frame (see (\ref{eq:transform})). It is seen that points near to the center of the image after projection (i.e., small $X_i/Z_i, Y_i/Z_i$) lead to small Jacobian. Therefore it is beneficial to have edge features equally distributed in the image. Moreover, since LiDAR  noises increase with distance as in (\ref{eq:meas_noise_cov}), the calibration scene should have moderate depth. 
\subsection{\textcolor{black}{Initialization and Rough Calibration}}
\textcolor{black}{The presented optimization-based extrinsic calibration method aims for high-accuracy calibration but requires a good initial estimate of the extrinsic parameters that may not always be available. To widen its convergence basin, we further integrate an initialization phase into our calibration pipeline where the extrinsic value is roughly calibrated by maximizing the percent of edge correspondence ($P.C.$) defined below: 
\begin{equation}
    P.C. =\frac{N_{match}}{N_{sum}}
\end{equation}
where $N_{sum}$ is the total number of LiDAR edge points and $N_{match}$ is the number of matched LiDAR edge points. The matching is based on the distance and direction of a LiDAR edge point (after projected to the image plane) to its nearest edge in the image (see Section \ref{sec:matching}). The rough calibration is performed by an alternative grid search on rotation (grid size $0.5^{\circ}$) and translation (grid size $2$cm) over a given range. 
}

\vspace{-0.2cm}\section{Experiments and Results}
\begin{figure}[t]
    \vspace{-0.2cm}
    \centering
    \includegraphics[width=0.8\linewidth]{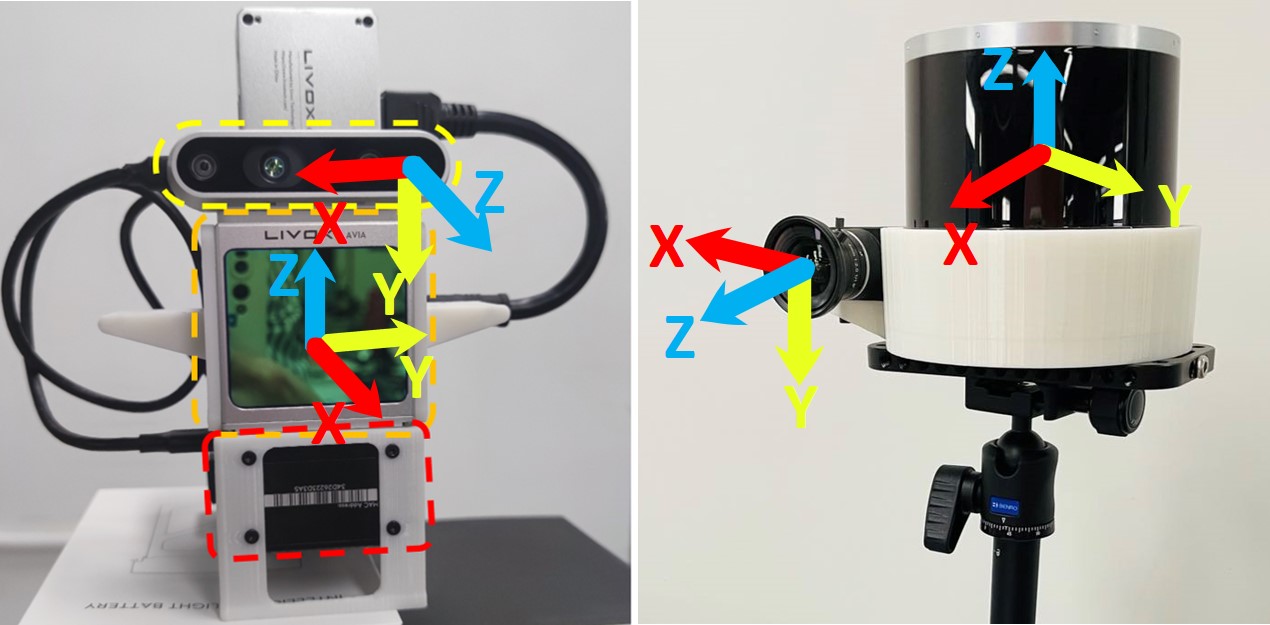}
    \caption{\textcolor{black}{Our sensor suite. Left is Livox Avia\protect\footnotemark[1] LiDAR and Intel Realsense-D435i\protect\footnotemark[2] camera, \textcolor{black}{which is used for the majority of the experiments (Section IV.A and B)}. Right is spinning LiDAR (OS2-64\protect\footnotemark[3] )and industry camera  (MV-CA013-21UC\protect\footnotemark[4]), \textcolor{black}{which is used for verification on fixed resolution LiDAR (Section IV.C)} . Each sensor suite has a nominal extrinsic, e.g., for Livox AVIA, it is $(0,-\frac{\pi}{2},\frac{\pi}{2}).$ for rotation (ZYX Euler Angle) and zeros for translation. }}
    \label{fig:sensor_suite}
    \vspace{-0.5cm}
\end{figure}

In this section, we validate our proposed methods in a variety of real-world experiments. We use a solid-state LiDAR called Livox AVIA, which achieves high-resolution point-cloud measurements at stationary due to its non-repetitive scanning \cite{livox}, and an Intel Realsense-D435i camera (see Fig.~\ref{fig:sensor_suite}). The camera intrinsic, including distortion models, has been calibrated beforehand. During data acquisition, we fix the LiDAR and camera in a stable position and collect point cloud and image simultaneously. The data acquisition time is 20 seconds for the Avia LiDAR to accumulate sufficient points. 

\footnotetext[1]{https://www.livoxtech.com/avia}
\footnotetext[2]{https://www.intelrealsense.com/depth-camera-d435i}
\footnotetext[3]{https://ouster.com/products/os2-lidar-sensor}
\footnotetext[4]{https://www.rmaelectronics.com/hikrobot-mv-ca013-21uc}


\vspace{-0.2cm}\subsection{Calibration Results in Outdoor and Indoor Scenes}
\begin{figure}[ht]
    \vspace{-0.2cm}
    \centering
    \includegraphics[width=1\linewidth]{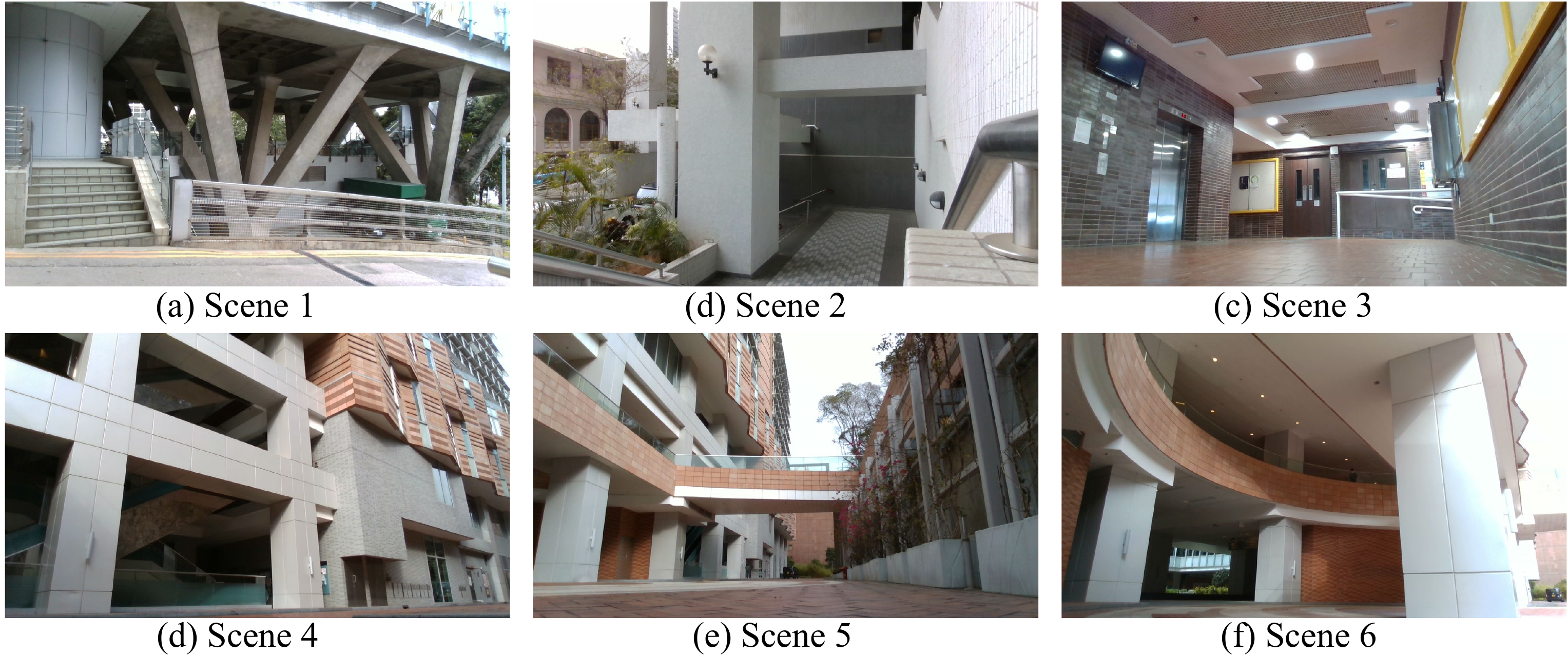}
    \centering
    \caption{Calibration scenes.}
    \label{fig:scenarios}
    \vspace{-0.6cm}
\end{figure}

We test our methods in a variety of indoor and outdoor scenes shown in Fig. \ref{fig:scenarios}, and validate the calibration performance as follows. 

\subsubsection{\textcolor{black}{Robustness and Convergence Validation}}\label{sec:convergence}
\begin{figure}[h]
    \vspace{-0.1cm}
    \centering
    \includegraphics[width=1\linewidth]{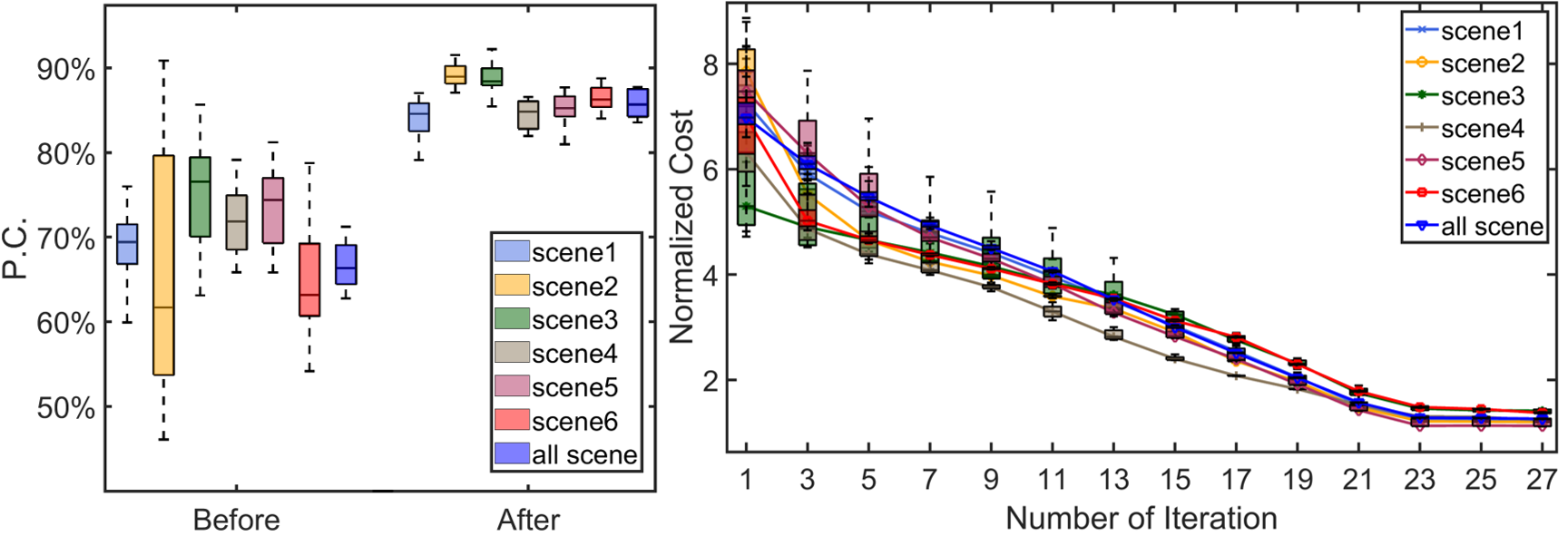}
    \caption{\textcolor{black}{Percent of edge correspondence before and after the rough calibration (left) and cost during the optimization (right).} }
    \label{fig:cost_value}
    \vspace{-0.2cm}
\end{figure}

\textcolor{black}{To verify the robustness of the full pipeline, we test it on each of the 6 scenes individually and also on all scenes together. For each scene-setting, 20 test runs are conducted, each with a random initial value uniformly drawn from a neighborhood ($\pm 5^{\circ}$ for rotation and $\pm 10$cm for translation) of the value obtained from the CAD model. Fig. \ref{fig:cost_value} shows the percent of edge correspondence before and after the rough calibration and the convergence of the normalized optimization cost $\frac{1}{N_{match}}\mathbf r^T \left( \mathbf J_{\mathbf w} \boldsymbol{\Sigma} \mathbf J_{\mathbf w} \right)^{-1} \mathbf r$ during the fine calibration. It is seen that, in all 7 scene settings and $20$ test runs of each, the pipeline converges for both rough and fine calibration.}

\begin{figure}[h]
    \vspace{-0.1cm}
    \centering
    \includegraphics[width=1\linewidth]{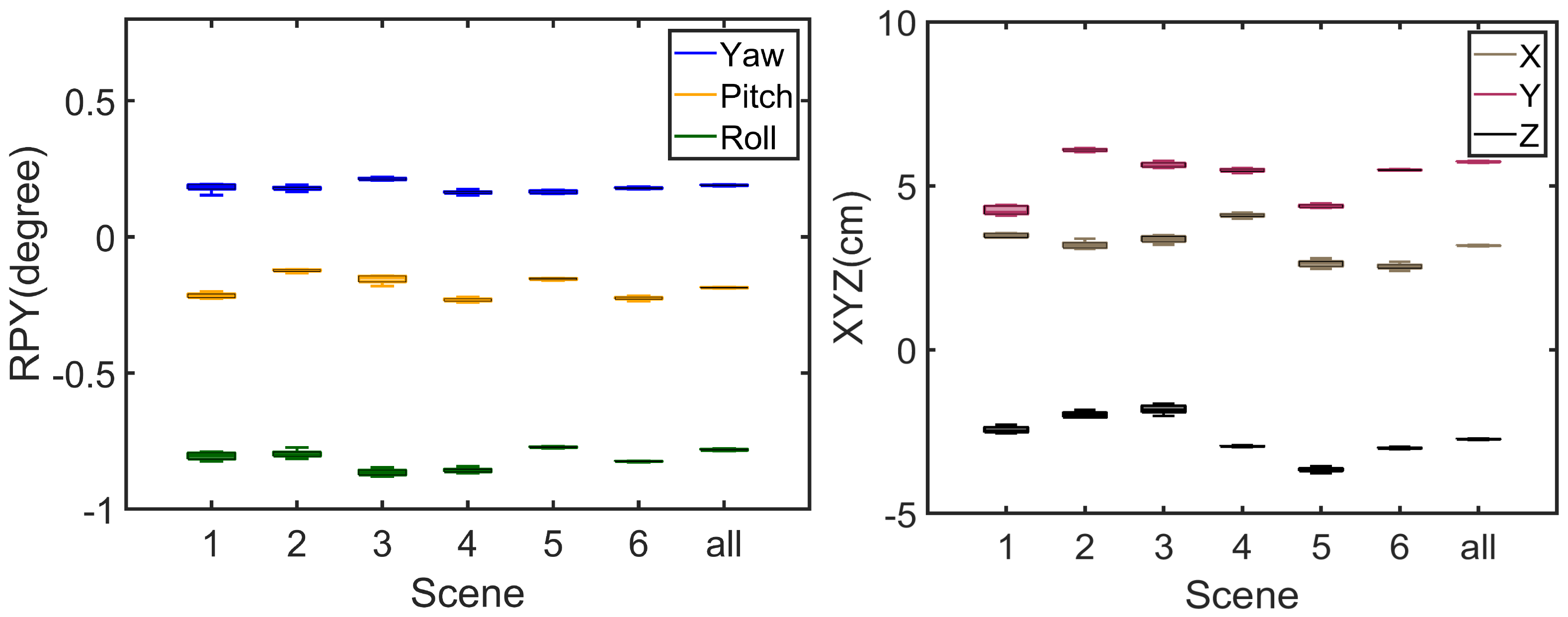}
    \caption{\textcolor{black}{Distribution of converged extrinsic values for all scene settings. The displayed extrinsic has its nominal part removed. }}
    \label{fig:extrinsic_distrubution}
    \vspace{-0.2cm}
\end{figure}

\begin{figure}
    \vspace{-0.3cm}
    \centering
    \includegraphics[width=1\linewidth]{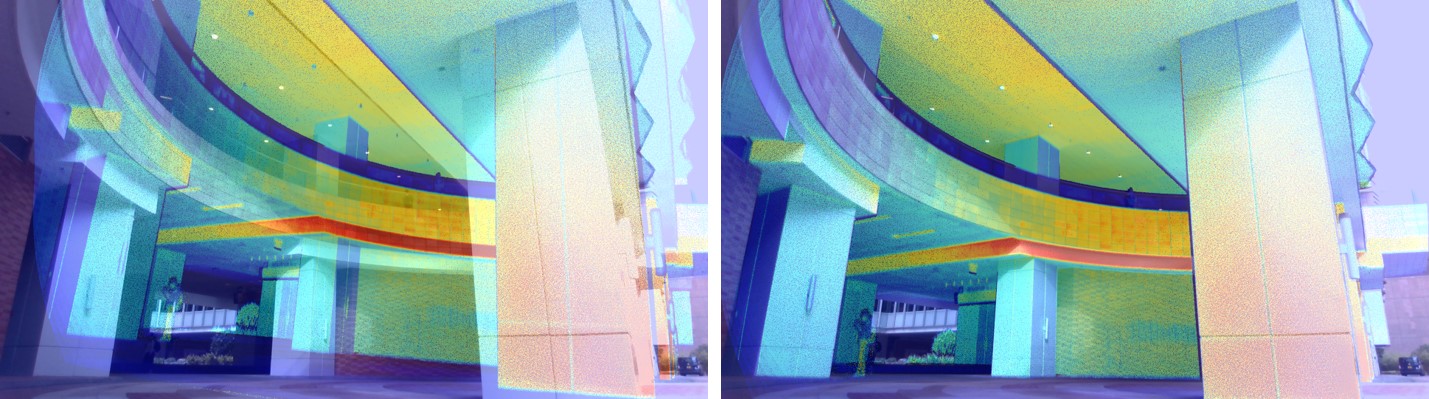}
    \caption{\textcolor{black}{LiDAR projection image overlaid on the camera image with an initial (left) and calibrated extrinsic (right). The LiDAR projection image is colored by Jet mapping on the measured point intensity.}}
    \label{fig:projection_validation}
    \vspace{-0.4cm}
\end{figure}

\textcolor{black}{Fig. \ref{fig:extrinsic_distrubution} shows the distribution of converged extrinsic values for all scene settings. It is seen that in each case, the extrinsic value converges to almost the same value regardless of the large range of initial value distribution. A visual example illustrating the difference before and after our calibration pipeline is shown in Fig. \ref{fig:projection_validation}. Our entire pipeline, including feature extraction, matching, rough calibration and fine calibration, takes less than 60 seconds}

\subsubsection{Consistency Evaluation}

\begin{figure}
    \vspace{-0.2cm}
    \centering
    \includegraphics[width=1\linewidth]{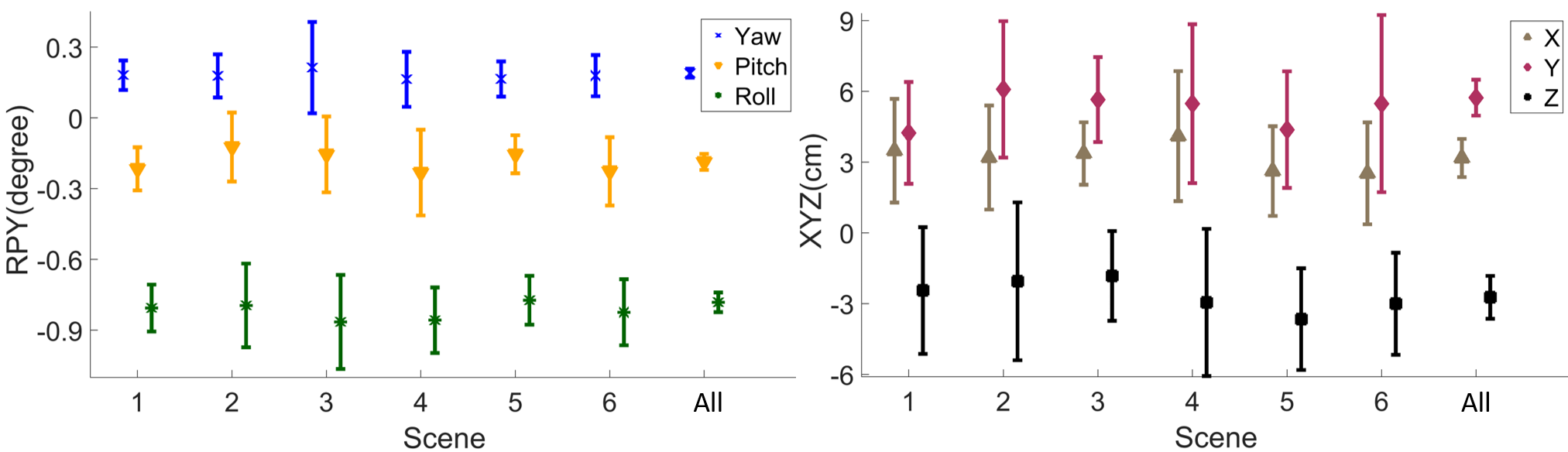}
    \caption{\textcolor{black}{Calibrated extrinsic with $3\sigma$ bound in all 7 scene settings.  The displayed extrinsic has its nominal part removed. }  }
    \label{fig:extrinsic and standard deviation}
    \vspace{-0.2cm}
\end{figure}

To evaluate the consistency of the extrinsic calibration in different scenes, we compute the standard deviation of each degree of freedom of the extrinsic:
\begin{equation}
\begin{aligned}
    \sigma_k&=\sqrt{ \left( \boldsymbol{\Sigma}_{\mathbf T} \right)_{k, k}},\ \ k\in \{ 1,2,...,6 \}
\end{aligned}
\label{standart deviation}
\end{equation}
where ${\Sigma}_{\mathbf T}$ is computed from (\ref{eq:estiamte_cov}). \textcolor{black}{As shown in Fig. \ref{fig:extrinsic_distrubution}, the converged extrinsic in a scene is almost identical regardless of its initial value, we can therefore use this common extrinsic to evaluate the standard deviation of the extrinsic calibrated in each scene. The results are summarized in Fig. \ref{fig:extrinsic and standard deviation}. It is seen that the $3\sigma$ of all scenes share overlaps and as expected, the calibration result based on all scenes have much smaller uncertainty and the corresponding extrinsic lies in the $3\sigma$ confidence level of the other 6 scenes. These results suggest that our estimated extrinsic and covariance are consistent. } 
\subsubsection{Cross Validation}
\begin{figure}
    \centering
    \includegraphics[width=1\linewidth]{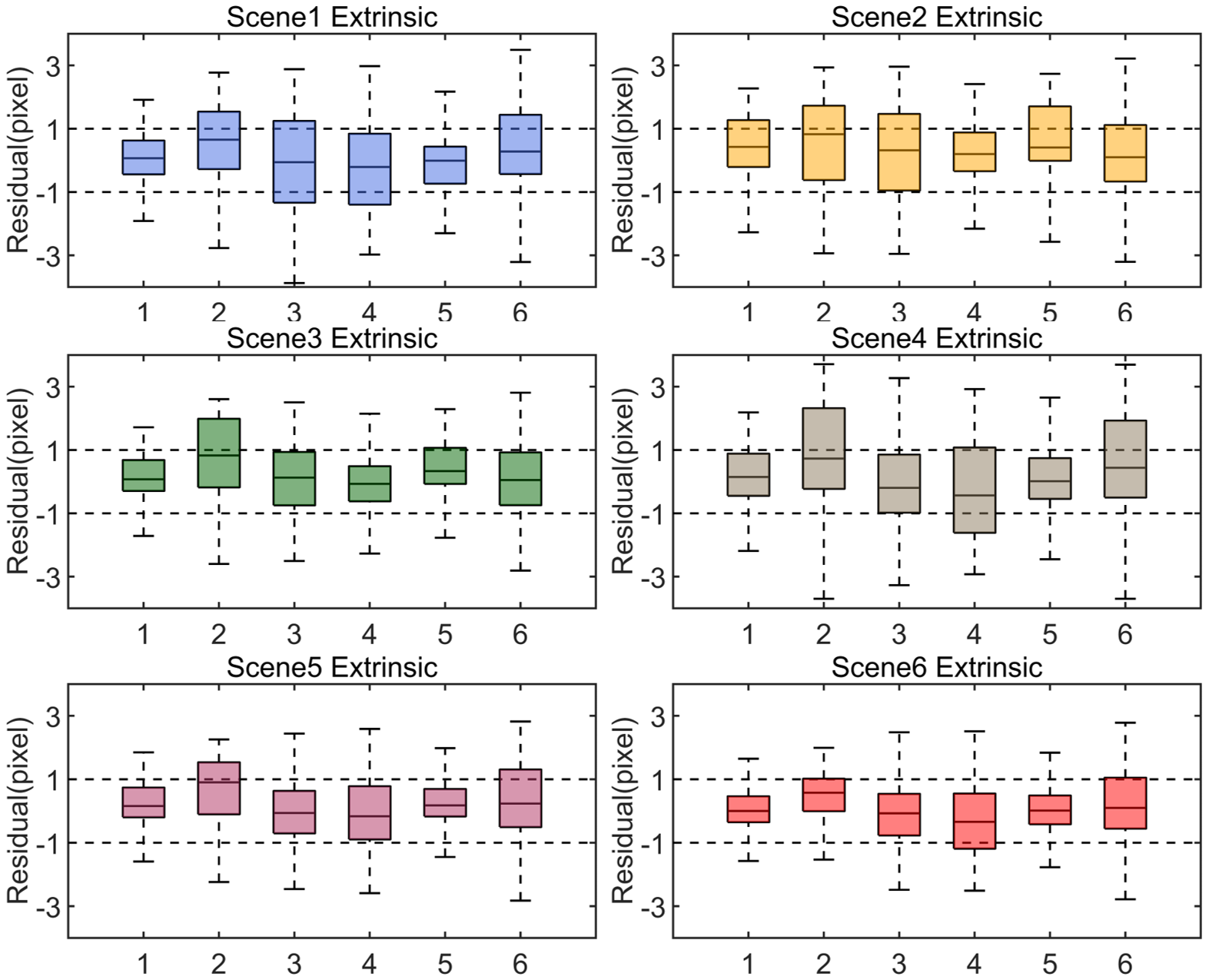}
    \caption{Cross validation results. The 6 box plots in $i$-th subfigure summarizes the statistical information (from up to down: maximum, third quartile, median, first quartile, and minimum) of residuals of the six scenes applied with extrinsic calibrated from scene $i$. }
    \label{fig:cross_validation}
    \vspace{-0.4cm}
\end{figure}

We evaluate the accuracy of our calibration via cross validation among the six individual scenes. To be more specific, we calibrate the extrinsic in one scene and apply the calibrated extrinsic to the calibration scene and all the rest five scenes. We obtain the residual vector $\mathbf r$ whose statistical information (e.g., mean, median) reveal the accuracy quantitatively. The results are summarized in Fig.~\ref{fig:cross_validation}, where the $20\%$ largest residuals are considered as outliers and removed from the figure. It is seen that in all calibration and validation scenes (36 cases in total), around $50\%$ of residuals, including the mean and median, are within one pixel. This validates the robustness and accuracy of our methods. 

\subsubsection{Bad Scenes}

\begin{figure}
    \centering
    \includegraphics[width=1\linewidth]{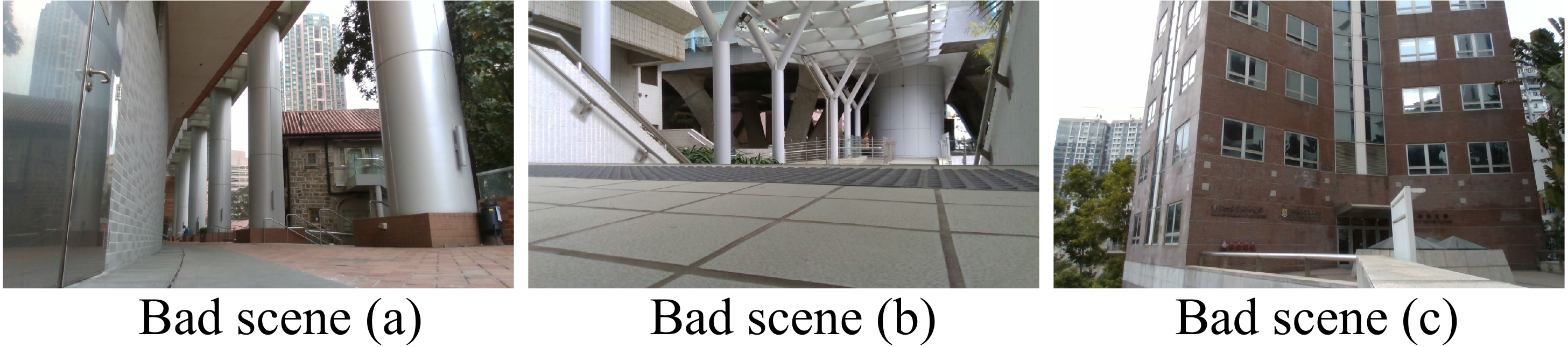}
    \caption{Examples of bad scenarios.}
    \label{fig:bad_scene}
    \vspace{-0.5cm}
\end{figure}

As analyzed in Section \ref{sec:analysis}, our method requires a sufficient number of edge features distributed properly. This does put certain requirements to the calibration scene. We summarize the scenarios in which our algorithm does not work well in Fig.\ref{fig:bad_scene}. The first is when the scene contains all cylindrical objects. Because the edge extraction is based on plane fitting, round objects will lead to inaccurate edge extraction. \textcolor{black}{Besides, cylindrical objects will also cause parallax issues, which will reduce calibration accuracy}. The second is the uneven distribution of edges in the scene. For example, most of the edges are distributed in the upper part of the image, which forms poor constraints that are easily affected by measurement noises. The third is when the scene contains edges only along one direction (e.g., vertical), in which the constraints are not sufficient to uniquely determine the extrinsic parameters.
\subsection{Comparison Experiments}

\begin{figure}[t]
    \centering
    \includegraphics[width=0.9\linewidth]{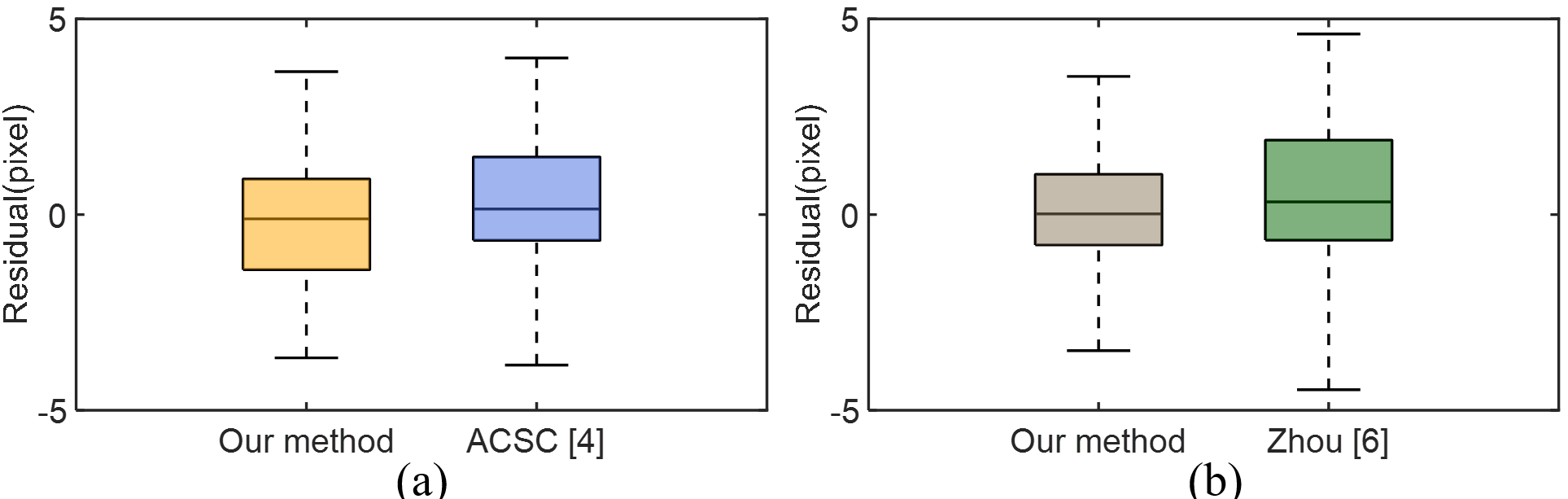}
    \caption{Comparison of residual distribution.}
    \label{fig:method_contrast}
    \vspace{-0.2cm}
\end{figure}

We compare our methods with \cite{ACSC, zhou2018automatic}, which both use checkerboard as a calibration target. The first one ACSC\cite{ACSC} uses point reflectivity measured by the LiDAR to estimate the 3D position of each grid corner on the checkerboard and computes the extrinsic by solving a 3D-2D PnP problem. Authors of \cite{ACSC} open sourced their codes and data collected on a Livox LiDAR similar to ours, so we apply our method to their data. Since each of their point cloud data only has 4 seconds data, it compromises the accuracy of the edge extraction in our method. To compensate for this, we use three (versus 12 used for ACSC \cite{ACSC}) sets of point clouds in the calibration to increase the number of edge features.
We compute the residuals in (\ref{eq:grad}) using the two calibrated extrinsic, the quantitative result is shown in {Fig. \ref{fig:method_contrast}} (a).   

\begin{figure}[t]
    \centering
    \includegraphics[width=0.9\linewidth]{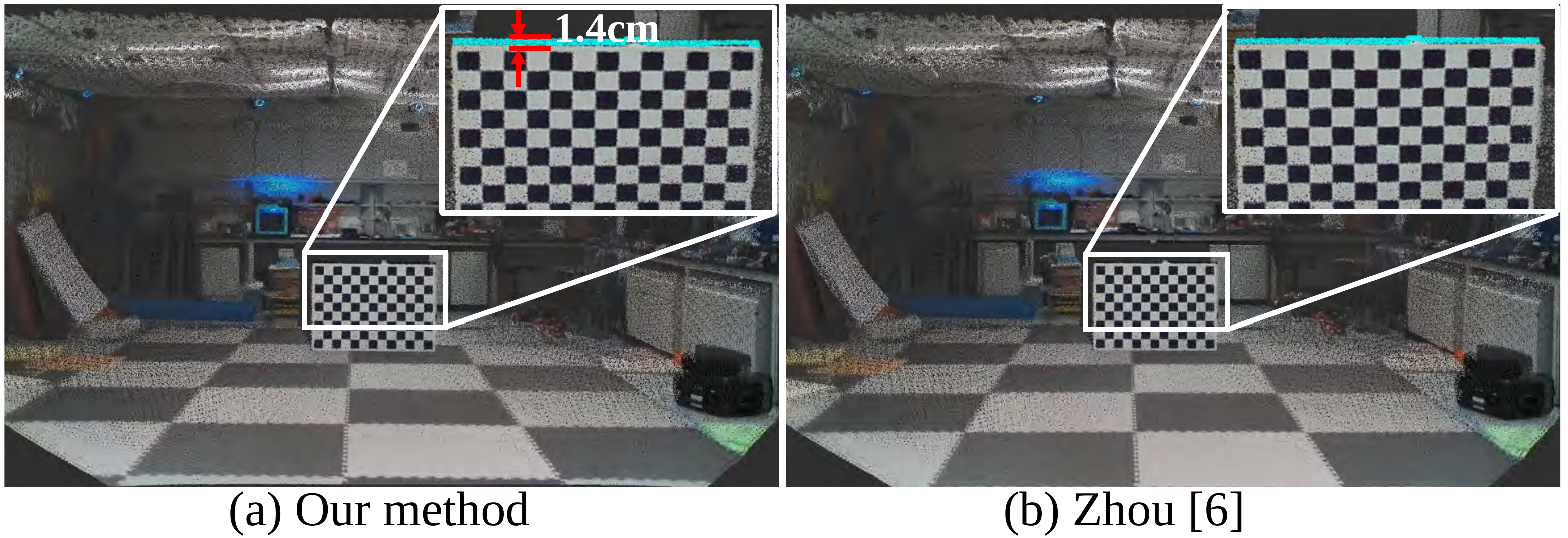}
    \caption{Colored point cloud: (a) our method; (b) Zhou \cite{zhou2018automatic}.}
    \label{fig:iros_contrast}
    \vspace{-0.6cm}
\end{figure}

The second method \cite{zhou2018automatic} estimates the checkerboard pose from the image by utilizing the known checkerboard pattern size. Then, the extrinsic is calibrated by minimizing the distance from LiDAR points (measured on the checkerboard) to the checkerboard plane estimated from the image. The method is designed for multi-line spinning LiDARs that have much lower resolution than ours, so we adopt this method to our LiDAR and test its effectiveness. The method in \cite{zhou2018automatic} also considers depth-discontinuous edge points, which are less accurate and unreliable on our LiDAR. So to make a fair comparison, we only use the plane points in the calibration when using \cite{zhou2018automatic}. To compensate for the reduced number of constraints, we place the checkerboard at more than 36 different locations and poses. In contrast, our method uses only one pair of data.  Fig. \ref{fig:iros_contrast} shows the comparison results on one of the calibration scenes with the  quantitative result supplied in {Fig. \ref{fig:method_contrast}} (b). It can be seen that our method achieves similar calibration accuracy with \cite{zhou2018automatic} although using no checkerboard information (e.g., pattern size). We also notice certain board inflation (the blue points in the zoomed image of Fig. \ref{fig:iros_contrast}), which is caused by laser beam divergence explained in Section \ref{sec:edge}. The inflated points 
are 1.4cm wide and at a distance of 6m, leading to an angle of ${0.014}/{6} = 0.13^\circ$, which agrees well with half of the vertical beam divergence angle (0.28\textdegree).

\subsection{\textcolor{black}{Applicability to Other Types of LiDARs}}
\textcolor{black}{Besides Livox Avia, our method could also be applied to conventional multi-line spinning LiDARs, which possess lower resolution at stationary due to the repetitive scanning. To boost the scan resolution, the LiDAR could be moved slightly (e.g., a small pitching movement) so that the gaps between scanning lines can be filled. A LiDAR (-inertial) odometry \cite{xu2021fast} could be used to track the LiDAR motion and register all points to its initial pose, leading to a much higher resolution scan enabling the use of our method. To verify this, We test our methods on another sensor platform (see Fig. \ref{fig:sensor_suite}) consisting of a spinning LiDAR (Ouster LiDAR OS2-64) and an industrial camera (MV-CA013-21UC).  The point cloud registration result and detailed quantitative calibration results are presented in the supplementary material ({https://github.com/ChongjianYUAN/SupplementaryMaterials}) due to the space limit. The results show that our method is able to converge to the same extrinsic for 20 initial values that are randomly sampled in the neighborhood ($\pm 3^{\circ}$ in rotation and $\pm 5$cm) of the value obtained from the CAD model.} 

\section{Conclusion}
This paper proposed a novel extrinsic calibration method for high resolution LiDAR and camera in targetless environments. We analyzed the reliability of different types of edges and edge extraction methods from the underlying LiDAR measuring principle. Based on this, we proposed an algorithm that can extract accurate and reliable LiDAR edges based on voxel cutting and plane fitting. Moreover, we theoretically analyzed the edge constraint and the effect of edge distribution on extrinsic calibration. Then a high-accuracy, consistent, automatic, and targetless calibration method is developed by incorporating accurate LiDAR noise models. Various outdoor and indoor experiments show that our algorithm can achieve pixel-level accuracy comparable to target-based methods. It also exhibits high robustness and consistency in a variety of natural scenes.

\bibliographystyle{unsrt}
\bibliography{reference}

\newcommand{\noop}[1]{}
\begin{thebibliography}{10}

\bibitem{xiao2018hybrid}
L.~Xiao, R.~Wang, B.~Dai, Y.~Fang, D.~Liu, and T.~Wu.
\newblock Hybrid conditional random field based camera-lidar fusion for road
  detection.
\newblock {\em Information Sciences}, 432:543–558, 2018.

\bibitem{di2019behavioral}
M.~Dimitrievski, P.~Veelaert, and W.~Philips.
\newblock Behavioral pedestrian tracking using a camera and lidar sensors on a
  moving vehicle.
\newblock {\em Sensors}, 19(2):391, 2019.

\bibitem{loamlivox}
J.~{Lin} and F.~{Zhang}.
\newblock Loam livox: A fast, robust, high-precision lidar odometry and mapping
  package for lidars of small fov.
\newblock In {\em 2020 IEEE International Conference on Robotics and Automation
  (ICRA)}, pages 3126--3131, 2020.

\bibitem{ACSC}
J.~Cui, J.~Niu, Z.~Ouyang, Y.~He, and D.~Liu.
\newblock Acsc: Automatic calibration for non-repetitive scanning solid-state
  lidar and camera systems, 2020.

\bibitem{koo2020analytic}
G.~Koo, J.~Kang, B.~Jang, and N.~Doh.
\newblock Analytic plane covariances construction for precise planarity-based
  extrinsic calibration of camera and lidar.
\newblock {\em 2020 IEEE International Conference on Robotics and Automation
  (ICRA)}, 2020.

\bibitem{zhou2018automatic}
L.~Zhou, Z.~Li, and M.~Kaess.
\newblock Automatic extrinsic calibration of a camera and a 3d lidar using line
  and plane correspondences.
\newblock {\em 2018 IEEE/RSJ International Conference on Intelligent Robots and
  Systems (IROS)}, 2018.

\bibitem{chen2020novel}
S.~Chen, J.~Liu, X.~Liang, S.~Zhang, J.~Hyyppa, and R.~Chen.
\newblock A novel calibration method between a camera and a 3d lidar with
  infrared images.
\newblock {\em 2020 IEEE International Conference on Robotics and Automation
  (ICRA)}, 2020.

\bibitem{livox}
Z.~Liu, F.~Zhang, and X.~Hong.
\newblock Low-cost retina-like robotic lidars based on incommensurable
  scanning.
\newblock {\em IEEE/ASME Transactions on Mechatronics}, page 1–1, 2021.

\bibitem{kummerle2020}
J.~Kummerle and T.~Kuhner.
\newblock Unified intrinsic and extrinsic camera and lidar calibration under
  uncertainties.
\newblock {\em 2020 IEEE International Conference on Robotics and Automation
  (ICRA)}, 2020.

\bibitem{gong2013}
X.~Gong, Y.~Lin, and J.~Liu.
\newblock 3d lidar-camera extrinsic calibration using an arbitrary trihedron.
\newblock {\em Sensors}, 13(2):1902–1918, 2013.

\bibitem{park2014}
Y.~Park, S.~Yun, C.~Won, K.~Cho, K.~Um, and S.~Sim.
\newblock Calibration between color camera and 3d lidar instruments with a
  polygonal planar board.
\newblock {\em Sensors}, 14(3):5333–5353, 2014.

\bibitem{zhu2020camvox}
Y.~Zhu, C.~Zheng, C.~Yuan, X.~Huang, and X.~Hong.
\newblock Camvox: A low-cost and accurate lidar-assisted visual slam system.
\newblock {\em arXiv preprint arXiv:2011.11357}, 2020.

\bibitem{pandey2012automatic}
G.~Pandey, J.~McBride, S.~Savarese, and R.~Eustice.
\newblock Automatic targetless extrinsic calibration of a 3d lidar and camera
  by maximizing mutual information.
\newblock In {\em Proceedings of the AAAI Conference on Artificial
  Intelligence}, volume~26, 2012.

\bibitem{scaramuzza2007extrinsic}
D.~Scaramuzza, A.~Harati, and R.~Siegwart.
\newblock Extrinsic self calibration of a camera and a 3d laser range finder
  from natural scenes.
\newblock In {\em 2007 IEEE/RSJ International Conference on Intelligent Robots
  and Systems}, pages 4164--4169. IEEE, 2007.

\bibitem{levinson2013automatic}
J.~Levinson and S.~Thrun.
\newblock Automatic online calibration of cameras and lasers.
\newblock In {\em Robotics: Science and Systems}, volume~2, page~7. Citeseer,
  2013.

\bibitem{nagy2019}
B.~Nagy, L.~Kovács, and C.~Benedek.
\newblock Online targetless end-to-end camera-lidar self-calibration.
\newblock In {\em 2019 16th International Conference on Machine Vision
  Applications (MVA)}, pages 1--6, 2019.

\bibitem{lidarcalib}
X.~Liu and F.~Zhang.
\newblock Extrinsic calibration of multiple lidars of small fov in targetless
  environments.
\newblock {\em IEEE Robotics and Automation Letters}, 6(2):2036--2043, 2021.

\bibitem{liu2021balm}
Z.~Liu and F.~Zhang.
\newblock Balm: Bundle adjustment for lidar mapping.
\newblock {\em IEEE Robotics and Automation Letters}, 6(2):3184--3191, 2021.

\bibitem{canny}
J.~{Canny}.
\newblock A computational approach to edge detection.
\newblock {\em IEEE Transactions on Pattern Analysis and Machine Intelligence},
  PAMI-8(6):679--698, 1986.

\bibitem{he2021embedding}
D.~He, W.~Xu, and F.~Zhang.
\newblock Embedding manifold structures into kalman filters.
\newblock {\em arXiv preprint arXiv:2102.03804}, 2021.

\bibitem{hertzberg2013integrating}
C.~Hertzberg, R.~Wagner, U.~Frese, and L.~Schr{\"o}der.
\newblock Integrating generic sensor fusion algorithms with sound state
  representations through encapsulation of manifolds.
\newblock {\em Information Fusion}, 14(1):57--77, 2013.

\bibitem{xu2021fast}
W.~Xu and F.~Zhang.
\newblock Fast-lio: A fast, robust lidar-inertial odometry package by
  tightly-coupled iterated kalman filter.
\newblock {\em IEEE Robotics and Automation Letters}, 6(2):3317--3324, 2021.

\end{thebibliography}
\end{document}